\documentclass[runningheads]{llncs}

\usepackage{eccv}

\usepackage{eccvabbrv}

\usepackage{graphicx}
\usepackage{booktabs}

\usepackage[accsupp]{axessibility}  

\usepackage{hyperref}

\usepackage{orcidlink}

\usepackage{comment}
\usepackage{tikz}
\usetikzlibrary{positioning, shapes.multipart}
\usepackage{placeins}
\usepackage{multirow}
\usepackage{wrapfig}
\usepackage{array}

\newcommand{\bftab}{\fontseries{b}\selectfont}

\newif\ifdraft
\drafttrue

\newcommand{\tg}[1]{\ifdraft{\textcolor{orange!60!black}{#1}}\else{#1}\fi}

\newcommand{\ouruparrow}{(\uparrow)}
\newcommand{\ourdownarrow}{(\downarrow)}

\newcommand{\tabhl}[1]{\bftab{#1}}

\DeclareRobustCommand{\citeNHL}[1]{%
  \begingroup
    \begin{NoHyper}\cite{#1}\end{NoHyper}%
  \endgroup
}

\newcommand{\samethanks}{\unskip${}^\star$}

\begin{document}

\title{Predictive Photometric Uncertainty in Gaussian Splatting for Novel View Synthesis} 

\titlerunning{3DGS-Uncertainty}

\author{Chamuditha Jayanga Galappaththige\inst{1,2}\thanks{ Equally contributed.} \and
Thomas Gottwald\inst{3}\samethanks \and
Peter Stehr\inst{3}\samethanks \and 
Edgar Heinert\inst{4} \and 
Niko Suenderhauf\inst{1,2} \and 
Dimity Miller\inst{1,2} \and  
Matthias Rottmann\inst{4}}
\authorrunning{Galappaththige et al.}

\institute{QUT Centre for Robotics, Australia \and
ARIAM Hub, Australia \and
University of Wuppertal, Germany \and University of Osnabrück, Germany}

\maketitle
\begin{abstract}

 Recent advances in 3D Gaussian Splatting have enabled impressive photorealistic novel view synthesis. However, to transition from a pure rendering engine to a reliable spatial map for autonomous agents and safety-critical applications, knowing where the representation is uncertain is as important as the rendering fidelity itself. We bridge this critical gap by introducing a lightweight, plug-and-play framework for pixel-wise, view-dependent predictive uncertainty estimation. Our post-hoc method formulates uncertainty as a Bayesian-regularized linear least-squares optimization over reconstruction residuals. This architecture-agnostic approach extracts a per-primitive uncertainty channel without modifying the underlying scene representation or degrading baseline visual fidelity. Crucially, we demonstrate that providing this actionable reliability signal successfully translates 3D Gaussian splatting into a trustworthy spatial map, further improving state-of-the-art performance across three critical downstream perception tasks: active view selection, pose-agnostic scene change detection, and pose-agnostic anomaly detection. Code is available at \href{https://chumsy0725.github.io/3DGS-Uncertainty/}{github.io/3DGS-Uncertainty}.


\end{abstract}

\section{Introduction}
\label{sec:intro}

Radiance fields~\cite{kerbl20233dgaussiansplatting,mildenhall2021nerf} have evolved from novel view synthesis (NVS) engines into foundational spatial maps for autonomous agents~\cite{matsuki2024gaussian, zhu2025loopsplat, fei20243d}. However, constructing a radiance field from 2D images is an inherently ill-posed inverse problem~\cite{tarantola2005inverse}. To function reliably in real-world deployments, where severe occlusions, unobserved regions, and geometric ambiguities are inevitable, these systems must be capable of rigorously quantifying their predictive uncertainty~\cite{sunderhauf2018limits}. 3D Gaussian Splatting (3DGS)~\cite{kerbl20233dgaussiansplatting} has rapidly emerged as the leading representation for these tasks, combining the expressiveness of volumetric rendering with the efficiency of rasterization. While recent advancements have drastically improved 3DGS visual fidelity~\cite{kheradmand20243d,ye2024absgs}, geometric consistency~\cite{chung2024depth,kerbl2024hierarchical}, and efficiency~\cite{hanson2025speedy,mallick2024taming}, equipping these models with robust, system-level uncertainty estimation (UE) remains a critical, underexplored challenge.

Existing UE methods for 3DGS fall into two complementary families, depending on whether they quantify uncertainty in the \emph{learned representation}, \ie, the Gaussian parameters, or in the \emph{rendered radiance field}, \ie, the rendered pixels. Most prior work targets the learned representation. Stochastic formulations~\cite{shen2021stochastic,aria2025modelinguncertainty,li2024variational,lyu2024manifold} model distributions over Gaussian parameters, but rely on sampling-based optimization or complex inference that introduces prohibitive latency, requires architectural modifications degrading baseline visual fidelity, and lacks modularity with the rapidly expanding ecosystem of 3DGS variants. Post-hoc alternatives instead estimate epistemic uncertainty in parameter space via Hessian approximations~\cite{jiang2024fisherrf,wilson2025popgs}. However, as shown in our experiments (Sec.~\ref{sec:NVS}), these parameter-centric methods capture view-dependent uncertainty poorly, hindering their use in downstream perception tasks.

\begin{figure}[t]
    \includegraphics[trim=0.0cm 0.0cm 1.4cm 3.3cm, clip, width=0.99\linewidth]{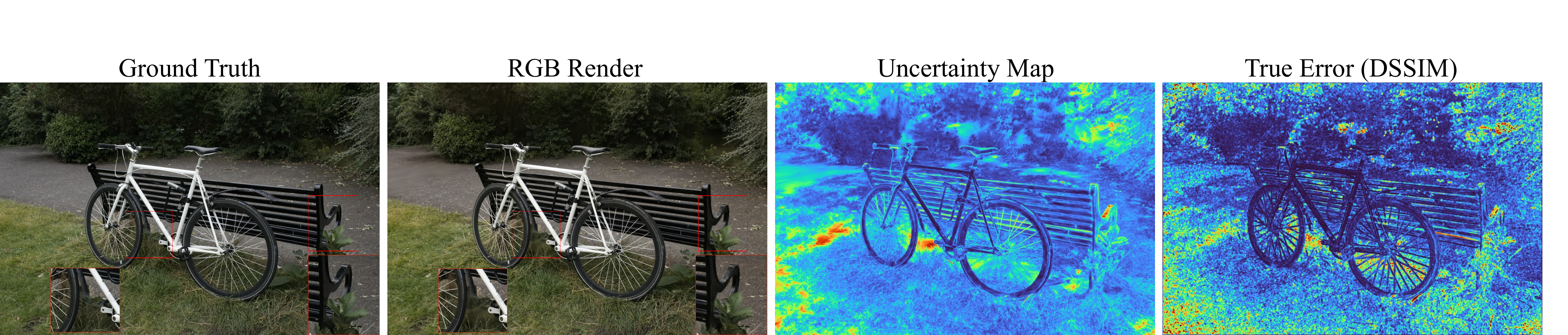}
    \captionof{figure}{Per-pixel UE for NVS. Our post-hoc method generates a view-dependent uncertainty map that closely mirrors regions of error within the RGB render.}
    \label{fig:hero}
\end{figure}

In contrast, we model uncertainty directly in the rendered radiance field~\cite{gottwald2025primu}, introducing an efficient, plug-and-play system that estimates pixel-wise \emph{predictive} uncertainty for NVS. Our approach generates view-dependent uncertainty maps alongside RGB images, with areas of high uncertainty reflecting regions of low rendering fidelity in the novel view (see Fig.~\ref{fig:hero}). This yields a single, actionable reliability signal that is particularly valuable for NVS, where downstream tasks consume and act directly upon the rendered pixels. We demonstrate this utility across three perception tasks, leveraging our uncertainty estimates to improve performance for next-best-view planning~\cite{pan2022activenerf}, pose-agnostic scene change detection~\cite{galappaththige2025multi}, and pose-agnostic anomaly detection~\cite{kruse2024splatpose}.

\begin{figure}[b]
    \centering
    \includegraphics[height=0.18\linewidth]{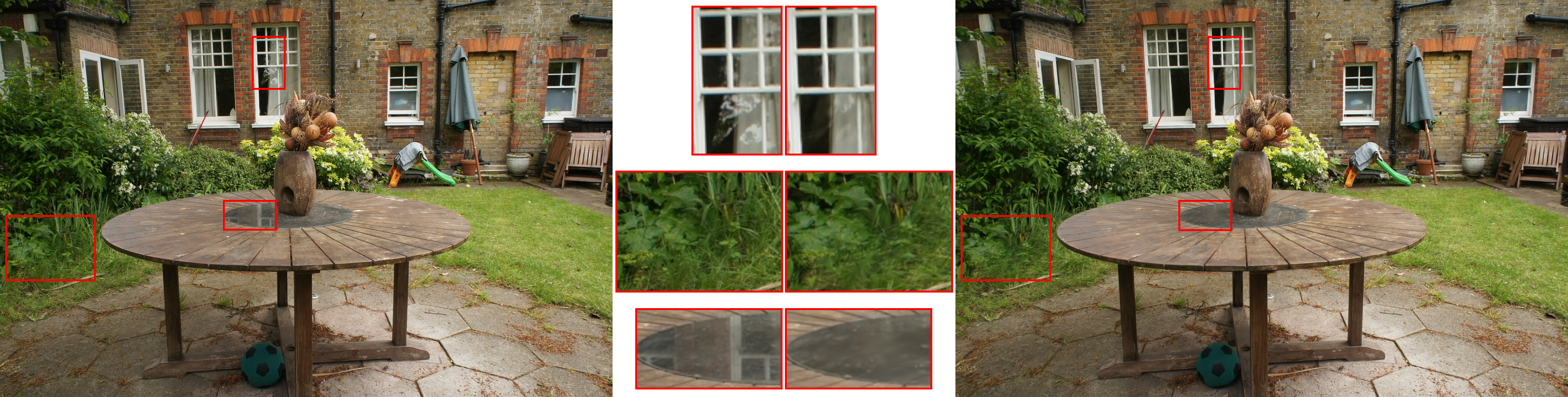}
    \includegraphics[height=0.18\linewidth]{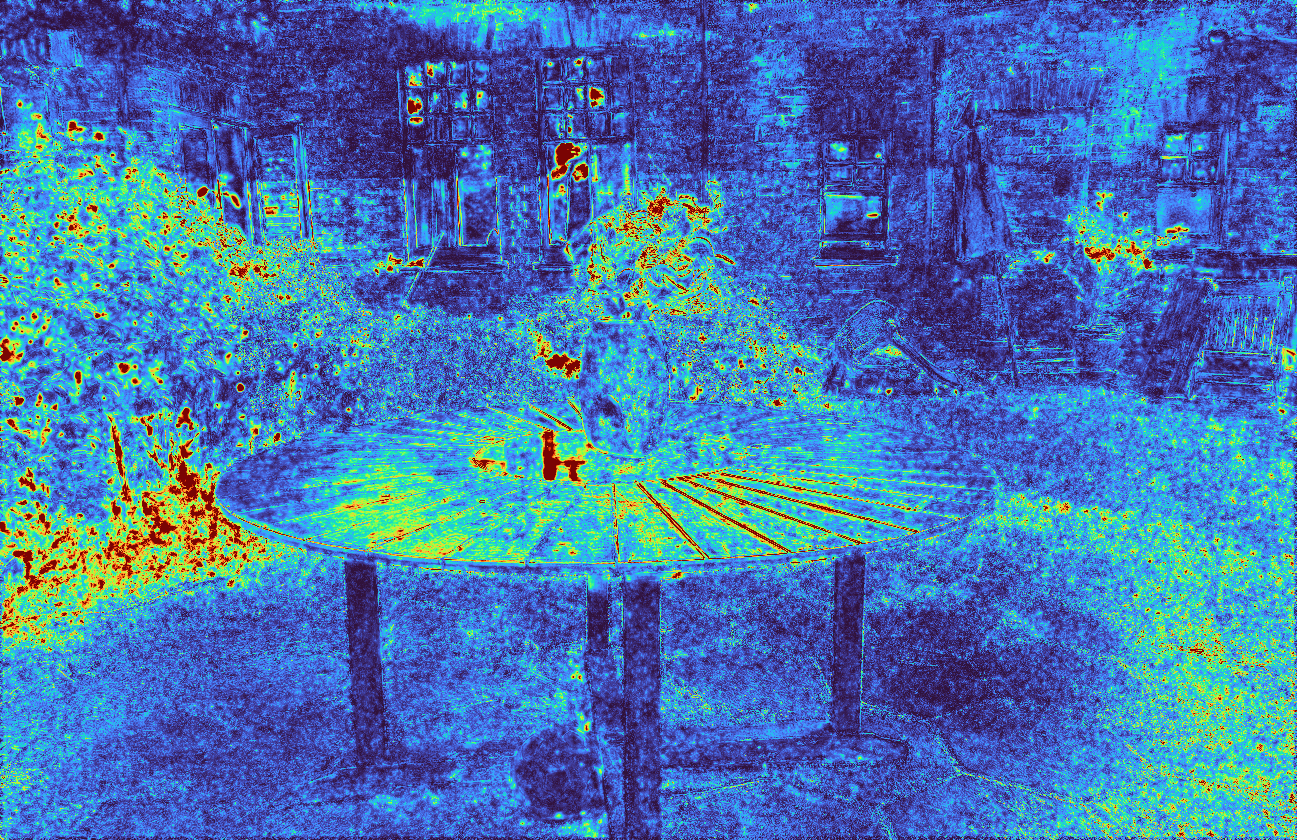}
    \caption{Visual example of reconstruction residual on training views. (Left) ground truth, (Center) rendering, (Right) residual map. The residual maps capture errors caused by both imperfect geometry (\eg, under-reconstructed regions in vegetation) and the limited capacity of view-dependent appearance (\eg, reflections on the table).}
    \label{fig:errorFactors}
\end{figure}

Our approach exploits a simple but powerful observation: regions that exhibit high reconstruction residual often correspond to areas where 3DGS fails to accurately represent the scene (see Fig.~\ref{fig:errorFactors}). 
We leverage this signal to guide the learning of a dedicated uncertainty channel within each 3DGS primitive, correlating representation error with high uncertainty.  
Yet reconstruction residuals alone are not a reliable proxy for uncertainty -- in sparse-view settings, the highly parameterized nature of 3DGS enables the model to overfit to the training views, driving residuals toward zero (thus signaling zero predictive uncertainty). To mitigate this, we introduce a Bayesian prior within our linear least-squares formulation of uncertainty estimation. This prior encourages uncertainty estimates to revert toward a maximal uncertainty level when insufficient multi-view evidence is available. This process is entirely post-hoc: it requires zero modifications to the base architecture or its optimization pipeline, ensuring preservation of the original rendering fidelity and compatibility with any 3DGS variant.  

In summary, we make the following claims:
\begin{itemize}
    \item We introduce a lightweight, plug-and-play system for view-dependent UE for novel view synthesis in 3DGS. We achieve state-of-the-art performance for UE while strictly preserving rendering fidelity (see Sec.~\ref{sec:NVS}).
    \item  We formulate this estimation as a (Bayesian-inspired) regularized linear least-squares problem, which ensures robust uncertainty predictions across both dense and sparse capture settings (see Sec.~\ref{sec:ue_on_highly_novel_views}).
    \item Beyond state-of-the-art UE in 3DGS, our uncertainty maps can be practically utilized in downstream tasks to improve state-of-the-art performance in active view selection (Sec.~\ref{sec:ue_avs}), pose-agnostic scene change detection (Sec.~\ref{sec:ue_scd}), and pose-agnostic anomaly detection (Sec.~\ref{sec:ue_pad}). 
\end{itemize}

\section{Related Work}

\subsection{Uncertainty Estimation in Gaussian Splatting}
\label{sec:RLGS}

UE in NeRFs~\cite{mildenhall2021nerf} has been extensively explored through stochastic sampling~\cite{seo2023flipnerf}, ensemble models~\cite{suenderhauf2023densityaware}, conditional normalizing flows~\cite{shen2022conditional}, and provenance modeling~\cite{nakayama2024provnerf}. Bayes' Rays~\cite{goli2024bayesrays} interprets uncertainty as allowable volumetric variation to identify under-constrained regions. While foundational, these designs remain tightly coupled to neural volumetric representations, making direct application to 3DGS~\cite{kerbl20233dgaussiansplatting} non-trivial due to its explicit, non-neural parameterization.

Existing UE techniques for 3DGS primarily rely on stochastic formulations. 
Continuous Semantic Splatting~\cite{wilson2024modelinguncertainty3dgaussian} learns semantic distributions over Gaussian primitives, interpreting their variance as a measure of semantic uncertainty. 
Several stochastic extensions~\cite{aria2025modelinguncertainty,li2024variational,lyu2024manifold} derive pixel-level uncertainty from multi-sample variance. 
Stochastic-GS~\cite{aria2025modelinguncertainty} learns probabilistic distributions over Gaussian parameters, while Variational-3DGS~\cite{li2024variational} leverages multi-scale Gaussian hierarchies to construct a diversified sampling space. 
Lyu \etal~\cite{lyu2024manifold} propose a low-dimensional manifold constraint to reduce the heavy sampling cost inherent in stochastic rendering.
Despite their advances, these stochastic methods are computationally expensive and demand substantial architectural or optimization modifications. This lacks plug-and-play applicability across 3DGS variants and 
often compromises
rendering fidelity.

Current post-hoc approaches for uncertainty estimation in 3DGS predominantly focus on the parameter space of the model. FisherRF~\cite{jiang2024fisherrf} uses Fisher information to quantify confidence in the learned representation. POp-GS~\cite{wilson2025popgs} extends this via optimal experimental design while also modeling inter-parameter correlations.
Similar to PRIMU~\cite{gottwald2025primu}, our approach operates directly in image space. However, rather than relying on hand-crafted features and a regressor trained on hold-out views, we introduce a lightweight linear least-squares formulation. This allows us to directly learn primitive-level, view-dependent uncertainty that efficiently aggregates into pixel-wise maps from arbitrary viewpoints, entirely bypassing the need for a regressor.

A parallel line of work assigns error-based scores to primitives to guide pruning and densification. PUP 3D-GS~\cite{hanson2025pup3dgs} and Speedy-Splat~\cite{hanson2025speedy} compute pruning scores based on $L^2$ loss sensitivity. Rota Bul\`o \etal~\cite{rota2025revising} derive a scalar densification score directly from the rendering error.
However, their per-pixel formulation ignores interactions between overlapping Gaussians, assigning elevated scalar scores to all contributing primitives even if only one is responsible for an artifact. While this heuristic suits densification, where over-densification is corrected in subsequent optimization, it falsely penalizes reliable primitives in frozen models. By analyzing all views simultaneously through our least-squares formulation, we disentangle these interactions to consistently identify correct Gaussians, even when they contribute to localized artifacts.

\subsection{Active View Selection in Gaussian Splatting}
\label{sec:RLAVS}

Active view selection (AVS) seeks the optimal next viewpoint to maximize reconstruction quality under a constrained acquisition budget~\cite{jiang2024fisherrf,pan2022activenerf,lyu2024manifold,suenderhauf2023densityaware,kopanas2023improvingnerf}. 
Standard approaches select a candidate view from a predefined pool to either minimize the overall uncertainty~\cite{suenderhauf2023densityaware,lyu2024manifold} or maximize the expected information gain~\cite{jiang2024fisherrf,pan2022activenerf}. 
Within 3DGS specifically, FisherRF~\cite{jiang2024fisherrf} optimizes information gain between candidate and training views, whereas Lyu \etal~\cite{lyu2024manifold} reduce reconstruction uncertainty via low-dimensional manifold sampling.
In contrast, we adopt a deterministic strategy that selects the next view to maximally reduce the per-primitive uncertainty estimated from our least-squares residual formulation, achieving state-of-the-art results in AVS.

\subsection{Pose-Agnostic Scene Change Detection}
\label{sec:RLSCD}

Scene change detection (SCD) has been widely studied as a bi-temporal comparison problem, where changes are identified between two aligned images~\cite{alcantarilla2018street,daudt2018fully,sakurada2015change,chen2021dr,varghese2018changenet,lei2020hierarchical,cyws2d,cyws3d,lin2025robust}. Such methods rely on strict viewpoint alignment and extensive supervision, limiting scalability in real-world applications such as robotic inspection~\cite{galappaththige2025multi}. 
Recent studies~\cite{kim2025towards,cho2025zero,kannanZero,alpherts2025emplace} explore zero-shot SCD using large visual foundation models~\cite{kirillov2023segment,oquab2023dinov2}; however, they still assume paired captures with minimal viewpoint disparity.

The emergence of high-fidelity 3D scene representations~\cite{kerbl20233dgaussiansplatting,wu20244d} has enabled \textit{pose-agnostic} SCD, where pre-change scenes are reconstructed and rendered into post-change viewpoints for comparison~\cite{galappaththige2025multi,lu20253dgs,jiang2025gaussian,galappaththige2025changes}. 
However, they remain highly sensitive to reconstruction inaccuracies and view-dependent artifacts, which are frequently misclassified as true scene changes. We address this critical flaw by leveraging our predicted per-pixel uncertainty to suppress unreliable regions, effectively filtering out rendering-induced false positives.
To the best of our knowledge, this is the first work to apply uncertainty estimation to pose-agnostic SCD. 
We further demonstrate that incorporating our uncertainty maps improves state-of-the-art pose-agnostic SCD methods~\cite{galappaththige2025multi}, as our UE accurately captures erroneous regions in novel-view renderings.

\subsection{Pose-Agnostic Anomaly Detection}
\label{sec:RLAD}

Traditional visual anomaly detection (AD) methods~\cite{Bergmann2019MVTecA, Roth2021TowardsTR} excel at localizing defects but typically require strictly controlled, fixed camera viewpoints. To relax this constraint, recent approaches leverage 3D scene representations~\cite{mildenhall2021nerf,kerbl20233dgaussiansplatting} to enable pose-agnostic AD~\cite{zhou2023pad,kruse2024splatpose,liu2024splatpose+}, allowing query captures from arbitrary viewpoints to be directly compared against novel-view renderings of a reconstructed reference model of an object. However, much like in pose-agnostic SCD, this direct comparison introduces a critical vulnerability: systemic rendering artifacts and geometric ambiguities are frequently misclassified as physical defects. By explicitly quantifying the reliability of the reference model, our view-dependent uncertainty maps provide a principled mechanism to mask these inherently ambiguous regions. This effectively suppresses rendering-induced false positives, significantly enhancing the robustness of the AD pipeline.

\section{Method}
\label{sec:method}

An overview of our approach is illustrated in \cref{fig:method}.  
The goal of this work is to predict \emph{pixel-wise, view-dependent uncertainty maps} for novel views while preserving the rendering fidelity of 3DGS.  
We begin by revisiting the fundamentals of volumetric rendering in 3DGS (\cref{sec:gs}) and reformulate the rendering process to predict uncertainty rather than color (\cref{sec:un}).  
Subsequently, we describe our learning formulation (\cref{sec:op}) and introduce a Bayesian-inspired regularization (\cref{sec:reg}) that improves uncertainty consistency in highly novel viewpoints.

\subsection{3D Gaussian Splatting}
\label{sec:gs}
In 3DGS~\cite{kerbl20233dgaussiansplatting}, a radiance field is represented using a set of 3D Gaussian primitives, each defined by a mean $\mu$, covariance matrix $\Sigma$, scalar opacity factor $a$, and view-dependent color $c(d)$ modeled via spherical harmonics (SH). Given a camera with position, viewing direction $d$, and intrinsic parameters, a view can be rendered by projecting the 3D Gaussians onto the image plane and $\alpha$-blending their contributions to obtain the final pixel color. This projection, also referred to as splatting~\cite{zwicker2001ewa}, employs a local affine approximation of the projective transformation, such that 3D Gaussians are mapped to 2D Gaussians on the image plane.
The opacity of a Gaussian at pixel $x$ is given by
$\alpha(x) = a \cdot \Phi (x, \mu^{\mathrm{2D}}, \Sigma^{\mathrm{2D}})$,
where $\mu^{\mathrm{2D}}$ and $\Sigma^{\mathrm{2D}}$ denote the mean and covariance of the projected 2D Gaussian, and $\Phi (x, \mu^{\mathrm{2D}}, \Sigma^{\mathrm{2D}})$ is its density at pixel $x$. The pixel-wise color is computed as a weighted sum over the primitives' color values along the viewing ray:
\begin{equation}
    C(x) = \sum_{k=1}^K c_k(d) \alpha_k(x) T_k(x),
\end{equation}
where $c_k(d)$ and $\alpha_k(x)$ denote the color and opacity of the $k$-th Gaussian, and $T_k(x) = \prod_{l=1}^{k-1} (1 - \alpha_l (x))$ is the transmittance up to the $(k-1)$-th Gaussian.

\subsection{Per-primitive Uncertainty}
\label{sec:un}
We estimate pixel-wise, view-dependent uncertainty by learning per-primitive uncertainty channels from training-view reconstruction residuals.
Since uncertainty may vary with the viewing direction, it is modeled using SH, analogously to color. For any view, a pixel-wise uncertainty map can be rendered in the same manner (and as fast) as an RGB image:
\begin{equation}\label{eq:UncertaintyRendering}
    U(x) = \sum_{k=1}^K u_k(d) \alpha_k(x) T_k(x),
\end{equation}
where $u_k(d)$ denotes the uncertainty of Gaussian $k$ from direction $d$.

\begin{figure}[t]
    \centering
    \resizebox{\textwidth}{!}{\input{Figs/method_graphic_downscaled/method_graphic}}
    \caption{
    Overview of our approach: We learn a primitive-level uncertainty channel $u_k$ on a \textit{well-trained} 3DGS, leveraging a (Bayesian-inspired) regularized linear least-squares formulation of training residuals. The learned view-dependent uncertainty can be rasterized to any given (novel) viewpoint $P_j$. Our method is post-hoc and therefore can be seamlessly integrated with any 3DGS variant without loss of fidelity.}
    \label{fig:method}
\end{figure}

We hypothesize that (in dense capture settings{\begingroup\renewcommand{\thefootnote}{\fnsymbol{footnote}}\footnote[2]{As in typical 3DGS datasets, \eg\ Mip-NeRF360.}\endgroup}) reconstruction residuals serve as a reliable proxy for \emph{predictive} uncertainty: just as 3DGS generalizes color to novel views, a well-fit residual channel approximates the expected photometric error there. This proxy aggregates aleatoric, epistemic, and optimization-related effects into a single, actionable signal.

While any discrepancy metric $L$ could serve as the target residual, we explicitly employ the same combination of $L^1$ loss and DSSIM used to train the base 3DGS model~\cite{kerbl20233dgaussiansplatting}.
To learn the primitive-level uncertainty channel $u_k$, we freeze all other representation parameters. This ensures the original reconstruction fidelity is perfectly preserved while the new uncertainty parameters are optimized to best explain the observed training errors.

We formulate the post-hoc estimation of the uncertainty channel as a least-squares problem. Given a well-reconstructed scene with $K$ primitives $G_1, \dotsc, G_K$, we enumerate all pixels across all training views as $x_1, \dotsc, x_M$, each corresponding to a viewing ray determined by its camera pose and pixel location. For simplicity, we initially assume $u_k$ to be view-independent and later describe how directional dependence can be incorporated. For each pixel $x_j$, depth-sorting the primitives front-to-back along its viewing ray defines a permutation $\sigma_j$, where $\sigma_j(k)$ denotes the index of the $k$-th primitive along the ray.
In this setting, \cref{eq:UncertaintyRendering} can be expressed as
\begin{equation}\label{eq:MatrixUncertaintyRendering}
    U(x_j) = \sum_{k=1}^K u_k \, \alpha_k(x_j) \, \hat{T}_k(x_j),
\end{equation}
where $\hat{T}_k(x_j) = \prod_{l=1}^{\sigma_j(k)-1} (1 - \alpha_{\sigma_j(l)}(x_j))$ and $u_k \in \mathbb{R}$ is the uncertainty of primitive $G_k$.
By gathering these volumetric blending weights into a matrix $A \in \mathbb{R}^{M \times K}$ such that
$A_{jk} = \alpha_k(x_j) \, \hat{T}_k(x_j)$, the rendered uncertainties for all pixels become the matrix-vector product $Au$, where $u = (u_1, \dotsc, u_K)^\top$. Let $y \in \mathbb{R}^{M}$ be the vector of corresponding pixel-wise reconstruction losses, such that $y_{j} = L_{x_j}$.
Estimating the primitive uncertainties that best explain these residuals amounts to solving the linear least-squares problem:
\begin{equation}\label{eq:least_squares}
    \arg\min_u \| y - A u \|_2^2.
\end{equation}
Incorporating view-dependency for $u$ via SH preserves this linearity; the SH basis functions are simply absorbed into A, and u expands to contain the concatenated SH coefficients across all primitives (see App. A for the explicit formulation).

\subsection{Optimization}
\label{sec:op}

Due to the massive scale of typical scenes, a direct solution to this problem is computationally intractable. Specifically, for the normal equations $A^TA u = A^T y$, the dimension of the matrix $A^T A$ scales with the number of primitives (often in the millions) times the 16 SH coefficients per Gaussian. Despite the sparsity of this matrix, its size renders direct solvers impractical.
Instead, we approximate the solution using stochastic gradient descent (SGD), following the same optimization procedure employed for the other Gaussian parameters. The convexity of the linear least-square objective ensures efficient and stable convergence.

After reconstruction is completed, for each pixel $x$ in the view, we first calculate the residual error $L_x$ 
for all training viewpoints. We adopt 3DGS's photometric error and keep $\lambda=0.2$ following 3DGS~\cite{kerbl20233dgaussiansplatting}: 
$L_x = (1 - \lambda) L^1_x + \lambda L^{\text{DSSIM}}_x$.
The overall training objective is then 
minimizing the $L^2$ loss of the residual $L_x$ and the rendered uncertainty $U_x$,
\begin{equation} \label{eq:training_objective}
    L = \sum_x (L_x - U_x)^2.
\end{equation}

Gradients are backpropagated exclusively to the uncertainty channels $u_k$, leaving the matrix $A$ fixed, thus ensuring linearity.

\subsection{Bayesian-Inspired Regularization}
\label{sec:reg}

The least-squares formulation in \cref{eq:least_squares} learns directional uncertainty only from viewing directions that are present in the training data. 
Consequently, directions absent from the training data provide no supervisory signal.
Furthermore, in sparse-view scenarios, a trivially low rendering error on the training views often masks a poorly reconstructed underlying 3D geometry.
As a result, uncertainty estimates for highly novel viewpoints can become unstable or unintuitive.

To this end, we introduce an $L^2$ regularization that imposes a prior over the directional uncertainty functions. In directions unsupported by training views, the learned uncertainty $u_k(d)$ is encouraged to default to a maximal uncertainty level $b \in \mathbb{R}$.
This approach is  grounded in the well-known interpretation of $L^2$-regularized linear regression as Bayesian inference with a Gaussian prior~\cite{bishop2006pattern} centered at the maximal uncertainty value $b$.

For a Gaussian $G_k$, the SH coefficients of its uncertainty feature define a function $u_k : S^2 \to \mathbb{R}$ on the unit sphere that encodes directional uncertainty. The difference of $u_k$ and the constant $b$
(which encodes maximal and isotropic uncertainty) is quantified by their $L^2$ distance:
\begin{equation} \label{eq:bayesian_loss_per_gaussian}
    l_k = \int_{S^2} (b - u_k(r))^2 \mathrm{d} r.
\end{equation}
This integral is approximated by evaluating $u_k(d)$ at discrete points on the sphere using Gauss-Legendre sampling~\cite{kowsky1986quadrature}.
Summing over all $K$ Gaussian primitives yields the global regularization loss:
$L_{\mathrm{reg}} = \sum_{k=1}^K l_k$.
The final training objective regularizes the base loss $L$ from \cref{eq:training_objective}\tg{:}
\begin{equation} \label{eq:bayesian_reg_loss}
    L^\prime = L + \lambda_{\mathrm{reg}} L_{\mathrm{reg}},
\end{equation}
where $\lambda_{\mathrm{reg}} \geq 0$ controls the influence of the prior.
The choice of $\lambda_{\mathrm{reg}}$ and conditions under which this regularization is beneficial are discussed in \cref{sec:ue_on_highly_novel_views}.
\section{Experiments}

\noindent\textbf{Datasets:}  
We follow the exact evaluation protocol established in 3DGS~\cite{kerbl20233dgaussiansplatting} and assess our UE performance on three standard benchmarks: 
Mip-NeRF360~\cite{barron2022mipnerf}, 
Tanks \& Temples~\cite{knapitsch2017tanksandtemples}, 
and Deep Blending~\cite{hedman2018deepblending}.

\noindent\textbf{Baselines:} 
We evaluate against representative state-of-the-art 3DGS UE approaches, including the post-hoc method FisherRF~\cite{jiang2024fisherrf} and the variational in\-ference-based methods of Lyu et al.~\cite{lyu2024manifold} (Manifold) and Li et al.~\cite{li2024variational} (Var3DGS). 
The latter two modify the 3DGS framework by learning distributions over Gaussian parameters, which slightly reduces rendering fidelity. 
A detailed comparison of their rendering quality against standard 3DGS is provided in App. D. 

\noindent\textbf{Metrics:}  
To evaluate how well predicted uncertainty maps correspond to actual reconstruction errors in NVS, we follow prior work~\cite{lyu2024manifold,jiang2024fisherrf,li2024variational} and report
Area Under the Sparsification Error curve (AUSE)~\cite{ilg2018uncertainty} and Pearson correlation. 

We report both metrics using two error formulations: the standard intensity difference $L^1$, and DSSIM. DSSIM is particularly important for evaluating errors in NVS, as these often arise from structural artifacts such as geometric misalignment, blur, or floaters rather than simple pixel-wise deviations. Since DSSIM is sensitive to luminance, contrast, and structural inconsistencies, it provides a more perceptually meaningful measure of NVS error than $L^1$.

\begin{figure}[t]
    \centering
    \includegraphics[width=0.99\linewidth]{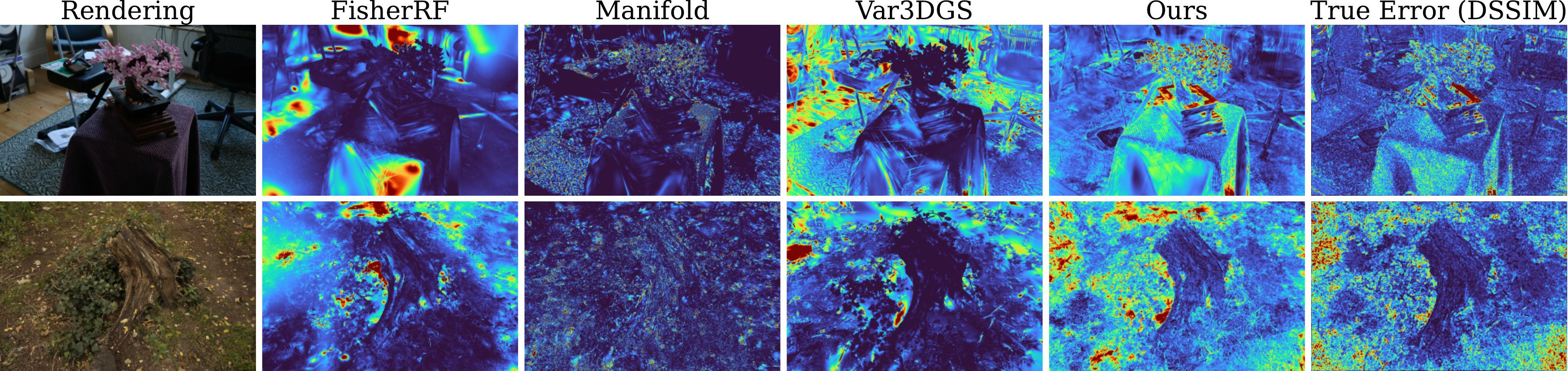}
    \caption{Qualitative comparison of predicted uncertainty maps on the \textit{bonsai} and \textit{stump} scenes of Mip-NeRF360~\cite{barron2022mipnerf}. Our UE more closely reflects the true error when compared to the baselines. Additional qualitative examples can be found in App. F.}
    \label{fig:qualitative_UE}
\end{figure}

\begin{table}[t]
    \centering
    \caption{Quantitative results for UE for NVS. We consistently outperform all baselines across all metrics and datasets while achieving minimal training overhead (OH).}
    \label{tab:ue_results}
    \setlength{\tabcolsep}{1.6pt} 
    \scalebox{0.70}{
    \begin{tabular}{@{}lrrrrcrrrrcrrrrc@{}}
        \toprule
        \multirow{3}{*}{\textbf{Method}} &
        \multicolumn{5}{c}{\textbf{Mip-NeRF360}~\cite{barron2022mipnerf}} &
        \multicolumn{5}{c}{\textbf{Tanks\&Temples}~\cite{knapitsch2017tanksandtemples}} &
        \multicolumn{5}{c}{\textbf{Deep Blending}~\cite{hedman2018deepblending}} \\
        \cmidrule(lr){2-6}\cmidrule(lr){7-11}\cmidrule(lr){12-16}
        & \multicolumn{2}{c}{AUSE$\ourdownarrow$} & \multicolumn{2}{c}{Pearson$\ouruparrow$} & \multirow{2}{*}{OH$\ourdownarrow$}
        & \multicolumn{2}{c}{AUSE$\ourdownarrow$} & \multicolumn{2}{c}{Pearson$\ouruparrow$} & \multirow{2}{*}{OH$\ourdownarrow$}
        & \multicolumn{2}{c}{AUSE$\ourdownarrow$} & \multicolumn{2}{c}{Pearson$\ouruparrow$} & \multirow{2}{*}{OH$\ourdownarrow$} \\
        \cmidrule(lr){2-3}\cmidrule(lr){4-5}\cmidrule(lr){7-8}\cmidrule(lr){9-10}\cmidrule(lr){12-13}\cmidrule(lr){14-15}
        & $L^1$ & DSSIM & $L^1$ & DSSIM & ($\%$) & $L^1$ & DSSIM & $L^1$ & DSSIM & ($\%$) & $L^1$ & DSSIM & $L^1$ & DSSIM & ($\%$) \\
        \midrule
        FisherRF~\cite{jiang2024fisherrf} & 0.708 & 0.606 & -0.055 & 0.009 & 14.2 & 0.691 & 0.709 & -0.087 & -0.145 & 19.3 & 0.751 & 0.853 & -0.116 & -0.190 & 19.7 \\
        Manifold~\cite{lyu2024manifold} & 0.520 & 0.559 & 0.070 & -0.005 & 30.2 & 0.574 & 0.654 & 0.053 & 0.008 & 23.6 & 0.503 & 0.548 & 0.095 & 0.074 & 48.1 \\
        Var3DGS~\cite{li2024variational} & 0.558 & 0.495 & 0.118 & 0.160 & $>$100 & 0.539 & 0.567 & 0.161 & 0.176 & $>$100 & 0.588 & 0.671 & 0.106 & 0.006 & $>$100 \\
        Ours & \tabhl{0.328} & \tabhl{0.214} & \tabhl{0.369} & \tabhl{0.547} & \tabhl{12.8} & \tabhl{0.299} & \tabhl{0.233} & \tabhl{0.427} & \tabhl{0.571} & \tabhl{14.0} & \tabhl{0.376} & \tabhl{0.356} & \tabhl{0.243} & \tabhl{0.244} & \tabhl{13.0} \\
        \bottomrule
    \end{tabular}}

\end{table}

\subsection{Uncertainty Estimation in NVS}
\label{sec:NVS}

\noindent\textbf{Analysis of results:} \cref{tab:ue_results} reports UE comparison on the hold-out (novel) views; here we set $\lambda_{\mathrm{reg}}=0$, as our Bayesian-inspired regularization yields only marginal gains in dense capture and is instead designed for sparse-view regimes (analyzed in \cref{sec:ue_on_highly_novel_views}). Our method consistently outperforms all baselines by a substantial margin. Notably, our method demonstrates particularly strong performance when evaluated using DSSIM error maps, indicating superior alignment with human perceptual error relative to existing methods. For instance, we achieve more than \emph{3$\times$ higher Pearson correlation} and \emph{less than half the AUSE} (DSSIM) compared to our best competitor Var3DGS~\cite{li2024variational} on Mip-NeRF360. Additionally, we achieve minimal training overhead (measured as percentage of additional time required for UE compared to vanilla 3DGS~\cite{kerbl20233dgaussiansplatting} training time).
Memory overhead is likewise minimal: we add only $(\text{SH degree}+1)^2$ scalars per Gaussian ($16$ at degree $3$), and freezing the base parameters keeps training memory below standard 3DGS.
App.~B further shows that our method is fully plug-and-play across different 3DGS variants, yielding similar or improved UE performance.
In \cref{fig:qualitative_UE}, our method is visually compared to the baselines, demonstrating a much higher correlation with the true error.

\noindent\textbf{Analysis on View-Dependence: }We ablate the view-dependence of our uncertainty channel $u_k$ by constraining the SH degree used to model it (\cref{tab:ue_view}). Similar to view-dependent color, the best performances are obtained with a degree of 3, while reducing the capacity degrades UE.

\begin{table}[t]
    \centering
    \caption{Results for (a) spherical harmonics degree ablation and (b) AVS.}
    \setlength{\tabcolsep}{4pt}
    \begin{subtable}[t]{0.48\textwidth}
        \centering
        \subcaption{Analysis of view-dependence on Mip-NeRF360~\cite{barron2022mipnerf}. Lowering the capacity of the spherical harmonic degrades the performance of our view-dependent uncertainty maps.}
        \label{tab:ue_view}
        \scalebox{0.75}{
        \begin{tabular}{@{}lrrrr@{}}
            \toprule
            \multirow{2}{*}{\textbf{Method}} &
            \multicolumn{2}{c}{AUSE $\ourdownarrow$} &
            \multicolumn{2}{c}{Pearson $\ouruparrow$} \\
            \cmidrule(lr){2-3} \cmidrule(lr){4-5}
            & L1 & DSSIM & L1 & DSSIM \\
            \midrule
            No View-Dependence & 0.391 & 0.310 & 0.217 & 0.356 \\
            SH Degree = 1      & 0.352  & 0.245 & 0.327 & 0.498 \\
            SH Degree = 2      & 0.331 & 0.218 & 0.353 & 0.535  \\
            SH Degree = 3      & \tabhl{0.328} & \tabhl{0.214}  & \tabhl{0.369} & \tabhl{0.547} \\
            \bottomrule
        \end{tabular}}
    \end{subtable}%
    ~
    \begin{subtable}[t]{0.48\textwidth}
        \centering
        \subcaption{Quantitative results for AVS on the Mip-NeRF360 dataset~\cite{barron2022mipnerf}. Our predicted uncertainty guided AVS achieves the best performance across all metrics.}
        \label{tab:avs_comparison}
        \scalebox{0.75}{
        \begin{tabular}{@{}lrrr@{}}
            \toprule
            \textbf{Method} & PSNR & SSIM & LPIPS\\
            & $\ouruparrow$ &
            $\ouruparrow$ & $\ourdownarrow$\\
            \midrule
            FisherRF~\cite{jiang2024fisherrf} & 20.266 & 0.593 & 0.363 \\
            Manifold~\cite{lyu2024manifold} & 19.732 & 0.595 & 0.373 \\
            Manifold$^{\dagger}$ & 20.088 & 0.611 & 0.350 \\
            Ours & \tabhl{20.676} & \tabhl{0.615} & \tabhl{0.344} \\
            \bottomrule
        \end{tabular}}
    \end{subtable}
\end{table}

\subsection{UE on Highly Novel Views}
\label{sec:ue_on_highly_novel_views}

\begin{figure}[b]
    \centering
    \begin{subfigure}[c]{0.46\textwidth}
        \centering
        \includegraphics[width=1\linewidth]{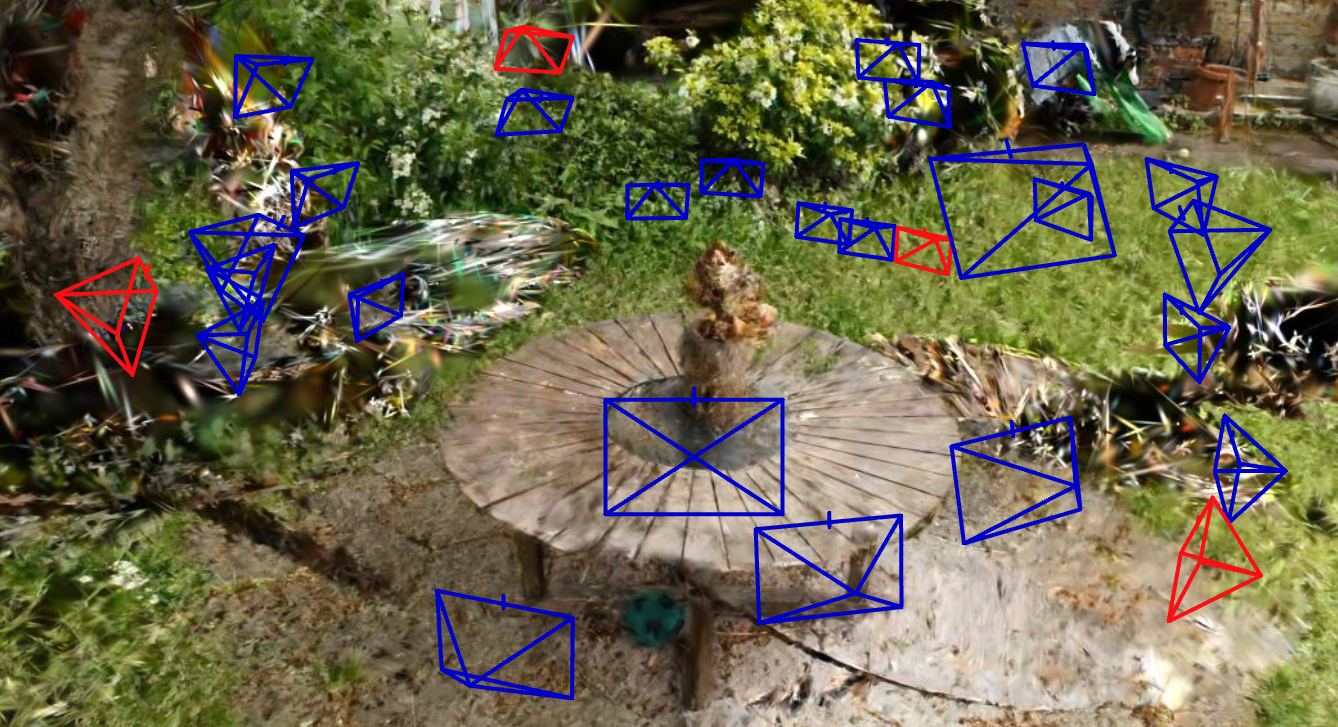}
        \caption{Visualization of the training and test split for UE on highly novel views. Red: training views; Blue: test views.}
        \label{fig:sparseUEsplit}
    \end{subfigure}%
    ~
    \begin{subfigure}[c]{0.505\textwidth}
        \centering
        \includegraphics[trim=0.7cm 0.82cm 0.81cm 0.72cm, clip, width=0.49\linewidth]{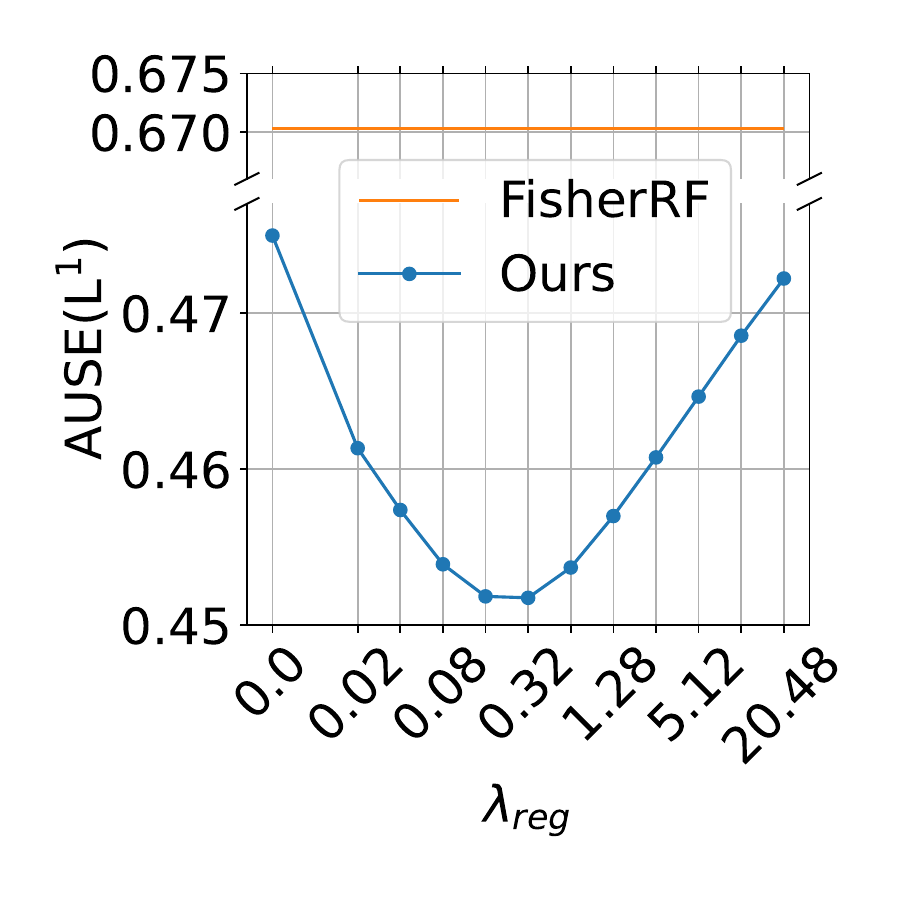}
        \includegraphics[trim=0.7cm 0.82cm 0.81cm 0.72cm, clip, width=0.49\linewidth]{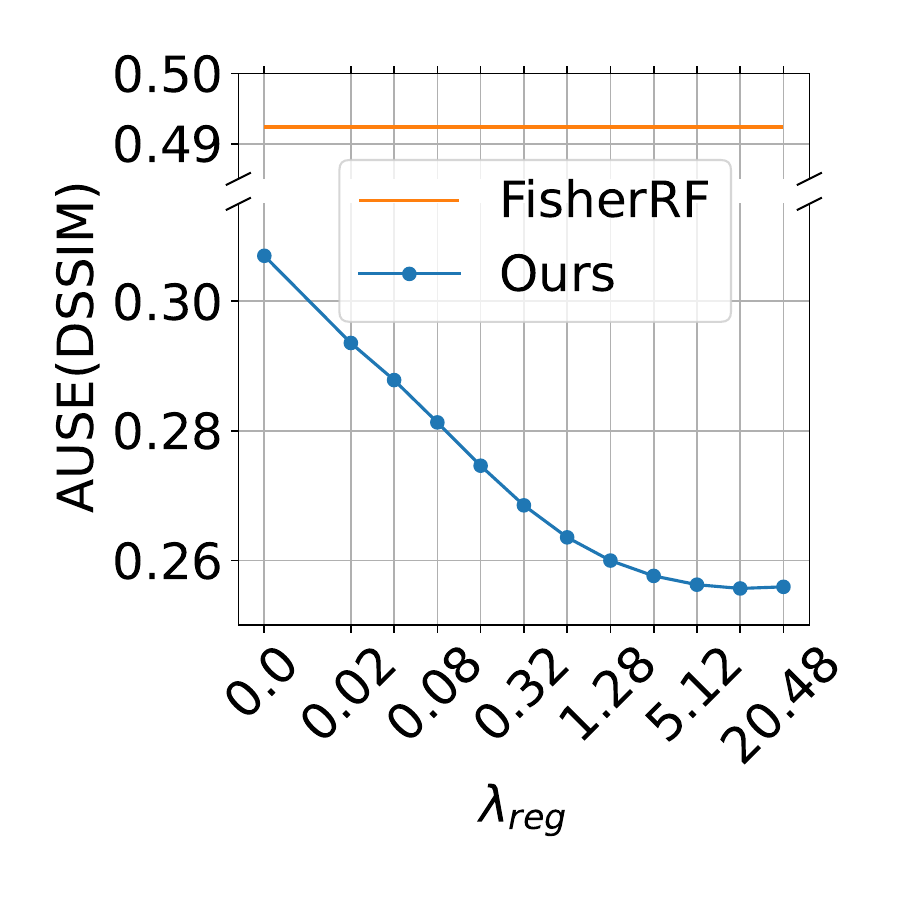}
        \caption{Quantitative results for UE on highly novel views on Mip-NeRF360~\cite{barron2022mipnerf} where our Bayesian-inspired regularization improves performance.}
        \label{fig:sparseUEplot}
    \end{subfigure}
    \caption{Highly novel view experiments: (a) example for view setup (b) results}
\end{figure}

\begin{figure}[t]
    \centering
    \includegraphics[width=1\linewidth]{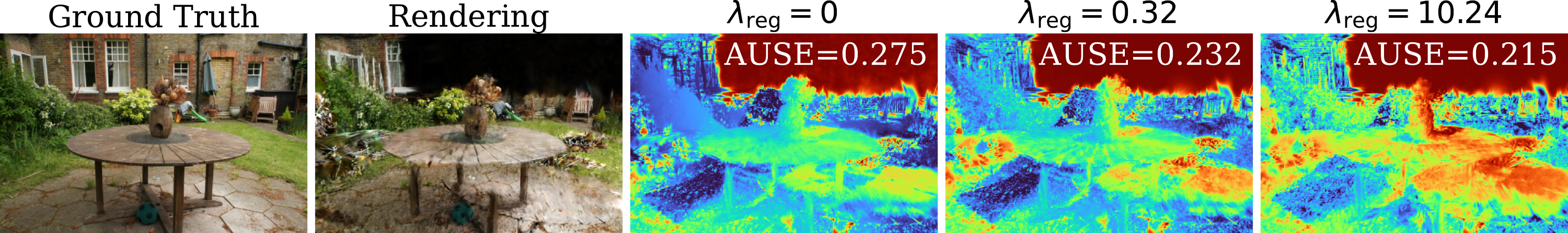}
    \caption{
    Qualitative examples of uncertainty maps with Bayesian-inspired regularization for different weights $\lambda_{\mathrm{reg}}$ (see \cref{eq:bayesian_reg_loss}) in the setting of highly novel views. The given AUSE values are with respect to DSSIM.
    }
    \label{fig:sparseUEexample}
\end{figure}

\noindent\textbf{Experiment Setup: }We investigate the contribution of our Bayesian-inspired regularization on highly novel views by adopting a sparse setup using four well-distributed training views. These four views are selected by maximizing pairwise camera center distances. We retain the test views in the standard setting (as in \cref{sec:NVS} and 3DGS~\cite{kerbl20233dgaussiansplatting}) for fair comparison. The train-test split is visualized in \cref{fig:sparseUEsplit}. Due to the sparse coverage of the scene, most test views observe the scene from highly novel viewing directions.

We evaluate on Mip-NeRF360~\cite{barron2022mipnerf}, training the base 3DGS model for 4,000 iterations and our uncertainty channel for 400 iterations to account for the sparse view count.
For experiments with Bayesian-inspired regularization, we set the maximal uncertainty level to $b = 1$. 
This choice follows naturally from our training objective, as uncertainty is regressed toward the photometric reconstruction residual, whose magnitude is bounded by the normalized image intensity range. 
Thus, a value of $1$ corresponds to maximal reconstruction error and provides a principled prior.
In these experiments, background regions are treated as maximally uncertain and assigned an uncertainty value of 1.
No explicit background prior is imposed when Bayesian regularization is disabled.
As a baseline, we compare against FisherRF~\cite{jiang2024fisherrf}, the only publicly available post-hoc method that can be directly applied to the same splatting outputs as ours.

\noindent\textbf{Analysis of Results: }In \cref{fig:sparseUEplot}, we study the effect of the regularization weight $\lambda_{\mathrm{reg}}$ by evaluating values from $0$ (no regularization) to $20.48$, doubling the weight from $0.02$ onward.
UE is assessed on both $L^1$ and DSSIM error maps.
Our approach consistently outperforms FisherRF across all $\lambda_{\mathrm{reg}}$ values in this challenging setting. Compared to the standard UE study, AUSE scores are generally higher, likely due to reduced splatting quality and fewer UE training views, which can make the residual error maps inconsistent and harder to learn. Overall, the Bayesian-inspired regularization improves performance in this regime, though its effect differs between error metrics. 
For $L^1$ error, the best AUSE scores are achieved for $0.16 \leq \lambda_{\mathrm{reg}} \leq 0.32$, while for DSSIM, optimal performance occurs at $\lambda_{\mathrm{reg}} = 10.24$, with marginal changes beyond.
Notably, these optimal ranges are largely scene-independent; scene-wise AUSE plots are provided in App. C.
A qualitative comparison of predicted uncertainty maps for different $\lambda_{\mathrm{reg}}$ values is shown in \cref{fig:sparseUEexample}.

\subsection{UE in Active View Selection}
\label{sec:ue_avs}
\noindent\textbf{Experiment Setup:}
We follow the AVS setup proposed by FisherRF~\cite{jiang2024fisherrf} and evaluate on the Mip-NeRF360 dataset~\cite{barron2022mipnerf}. Following prior work~\cite{jiang2024fisherrf, li2024variational, lyu2024manifold}, we report PSNR, SSIM, and LPIPS on the holdout views after reconstruction. Strictly adhering to the 3DGS evaluation protocol~\cite{kerbl20233dgaussiansplatting}, we evaluate at full resolution across all methods; while this naturally lowers metrics compared to downscaled versions in prior work~\cite{jiang2024fisherrf}, it guarantees a fair comparison.

We benchmark our approach against state-of-the-art AVS baselines FisherRF~\cite{jiang2024fisherrf} and the method by Lyu \etal~\cite{lyu2024manifold} (Manifold), which employs a stochastic variant of 3DGS. Because Manifold's variational optimization may degrade the visual fidelity of the underlying 3DGS representation, direct comparison is unfair. Therefore, we introduce Manifold$^{\dagger}$, a variant that uses Manifold's view predictions to guide a vanilla 3DGS~\cite{kerbl20233dgaussiansplatting}.

We initialize the AVS procedure with four training views selected by maximizing pairwise camera distances, consistent with our sparse-view setup. Each AVS method then sequentially selects 16 further views, resulting in a total of 20 training views. Between selections, the base 3DGS is trained for $100 \times N_\mathrm{views}$ iterations, where $N_\mathrm{views}$ is the current training view count. 
For ours, we train the uncertainty channel for $50 \times N_\mathrm{views}$ iterations (ablated in App.~E) prior to each selection, then selecting the candidate view with highest total uncertainty.

\noindent\textbf{Quantitative results:} As reported in \cref{tab:avs_comparison}, our method consistently outperforms all baselines across all metrics. This demonstrates that our predicted view-dependent uncertainty maps provide a highly reliable signal for identifying informative viewpoints, substantially improving downstream AVS performance.

\subsection{UE in Pose-Agnostic Scene Change Detection}
\label{sec:ue_scd}
\noindent\textbf{Experiment Setup:} Given multi-view captures of a pre-change (\textit{reference}) scene and a post-change (\textit{inference}) scene, pose-agnostic SCD first reconstructs a 3DGS representation of the reference scene \(\mathcal{R}_{\text{ref}}\) using reference images \(\mathcal{I}_{\text{ref}}\) and their SfM poses \(\mathcal{P}_{\text{ref}}\)~\cite{schonberger2016structure}. 
For each inference image \(I_{\text{inf}}^k\) with pose \(P_{\text{inf}}^k\), a corresponding rendered view \(I_{\text{ren}}^k\) is generated by querying \(\mathcal{R}_{\text{ref}}\), producing aligned image pairs \((I_{\text{ren}}^k, I_{\text{inf}}^k)\) for change detection.
Per-pixel change maps \(M^k\) are obtained in a label-free manner, either by comparing dense DINOv2 features~\cite{oquab2023dinov2}, direct pixel-level similarity~\cite{wang2004imagequality}, or a combination of both. A threshold is applied to produce binary masks \(M_{\text{bin}}^k\)~\cite{galappaththige2025multi}. 
The state-of-the-art MV3DCD~\cite{galappaththige2025multi} further aggregates the predictions \(\{M^k\}\) into a 3D change representation, improving robustness via multi-view consistency.

\begin{figure}[t]
    \centering
    \includegraphics[trim=0.0cm 0.0cm 21.2cm 0.0cm, clip, width=0.92\linewidth]{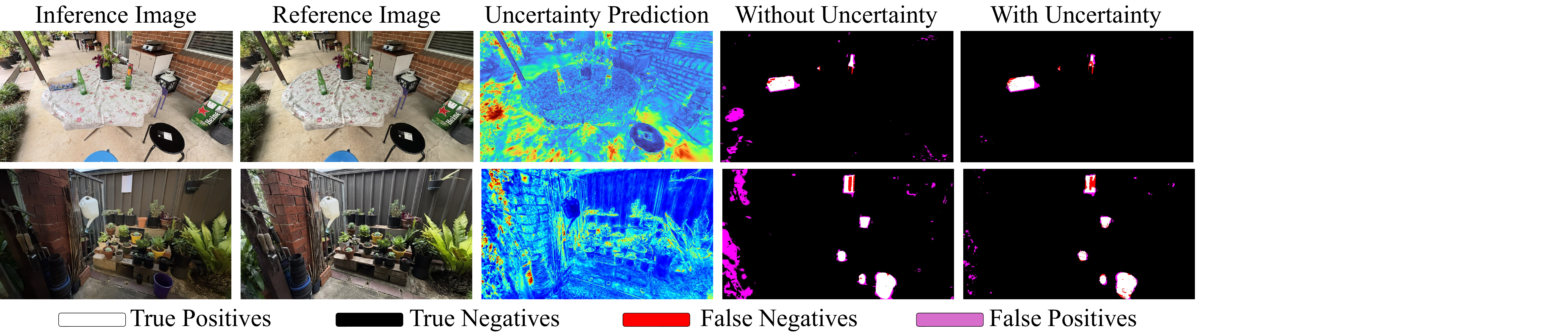}
    \caption{Qualitative examples of uncertainty-guided scene change detection. Our predicted uncertainty maps effectively capture rendering artifacts, allowing the system to suppress false positives caused by these artifacts rather than actual scene changes.}

    \label{fig:SCD_UE}
\end{figure}

However, the rendered views \(I_{\text{ren}}^k\) often contain artifacts (\cref{fig:SCD_UE}), caused by incomplete scene coverage, large capture pose discrepancies, or inherent 3DGS limitations. 
These inconsistencies lead to false positives, as differences between \(I_{\text{ren}}^k\) and \(I_{\text{inf}}^k\) may arise from rendering errors rather than actual scene changes.

\noindent\textbf{Leveraging uncertainty for pose-agnostic SCD:} 
We first apply our UE method from \cref{sec:method} to estimate uncertainties for the reference scene \(\mathcal{R}_{\text{ref}}\). 
Using these per-primitive values, we render an uncertainty map \(M^k_{\text{unc}}\) alongside the RGB image \(I_{\text{ren}}^k\) for each queried inference pose \(P_{\text{inf}}^k\). 
The standard change detection pipeline remains unchanged; however, before thresholding, we attenuate high-uncertainty pixels in the predicted change map \(M^k\), obtaining an uncertainty-guided change map \(\tilde{M}^k = M^k \odot (1 - M^k_{\text{unc}})\), where \(\odot\) denotes Hadamard multiplication. Following standard practice~\cite{galappaththige2025changes,galappaththige2025multi}, thresholding this map at the midpoint to obtain the final binary mask \(M_{\text{bin}}^k\) effectively filters out false positives induced by rendering artifacts rather than genuine scene changes.

\noindent\textbf{Quantitative Results:} 
We evaluate our approach on the PASLCD~\cite{galappaththige2025multi} benchmark following MV3DCD~\cite{galappaththige2025multi} and adopt four baselines for comparison. 
First, we include the \textit{Feature Diff.} baseline, which infers a change mask from differences between dense DINOv2~\cite{oquab2023dinov2} features. 
Second, we use MV3DCD’s zero-shot variant (MV3DCD-ZS), which fuses cues from feature- and structure-aware masks~\cite{galappaththige2025multi}. 
We further incorporate our uncertainty guidance into MV3DCD’s multi-view fusion framework and finally couple it with the existing state-of-the-art online pose-agnostic SCD approach, Online-SCD~\cite{galappaththige2025changes}.
As shown in \cref{tab:scd_comparison}, uncertainty-guidance consistently improves pose-agnostic SCD performance, yielding higher F1 scores and mean Intersection-over-Union (mIoU) across all baselines.

\subsection{UE in Pose-Agnostic Anomaly Detection}
\label{sec:ue_pad}
\noindent\textbf{Experiment Setup:} Given multi-view captures of a fault-free object, pose-agnostic AD first reconstructs a 3DGS reference model, denoted as $\mathcal{R}_{\text{ref}}$. During inference, for a given query image $I_{\text{inf}}^k$, its camera pose $P_{\text{inf}}^k$ is estimated on-the-fly~\cite{kruse2024splatpose,liu2024splatpose+}. Rendering $\mathcal{R}_{\text{ref}}$ at $P_{\text{inf}}^k$ yields a corresponding fault-free view, $I_{\text{ren}}^k$. The resulting aligned image pair $(I_{\text{ren}}^k, I_{\text{inf}}^k)$ is compared within the feature space of a pre-trained vision encoder~\cite{zhou2023pad} to generate a spatial anomaly score map $S^k$.

\noindent\textbf{Leveraging Uncertainty for Pose-Agnostic AD:} We apply our UE framework (described in \cref{sec:method}) to the optimized reference model $\mathcal{R}_{\text{ref}}$, rendering a per-pixel uncertainty map $M^k_{\text{unc}}$ along with $I_{\text{ren}}^k$. Crucially, we leave the underlying anomaly scoring pipeline untouched. Instead, we perform a post-hoc attenuation of the raw score map $S^k$ using our UE to produce an uncertainty-guided anomaly map: $\tilde{S}^k = S^k \odot (1 - M^k_{\text{unc}})$. This suppresses anomalous score activations in regions where the reference 3DGS model lacks confidence.

\noindent\textbf{Quantitative Results:} We evaluate our approach on the MAD-Real benchmark~\cite{zhou2023pad}, reporting AUROC and AUPRO metrics following prior work ~\cite{zhou2023pad,liu2024splatpose+,kruse2024splatpose}. By seamlessly integrating our uncertainty-guided attenuation into existing 3DGS-based AD baselines SplatPose~\cite{kruse2024splatpose} and SplatPosePlus~\cite{liu2024splatpose+}, we demonstrate a consistent improvement in AD across both architectures (see Tab.~\ref{tab:anomaly_detection}).

\begin{table}[t]
    \centering
    \caption{Numerical results for pose-agnostic SCD with uncertainty on PASLCD~\cite{galappaththige2025multi}. Our method consistently improves both state-of-the-art offline and online baselines. Percentage increase over the baseline is given in the $\Delta(\%)$ column.}
    \label{tab:scd_comparison}
    \setlength{\tabcolsep}{3pt}
    \scalebox{0.85}{ 
    \begin{tabular}{@{}lcccccccccccc@{}}
        \toprule
        & \multicolumn{3}{c}{\textbf{Feature Diff.}} & \multicolumn{3}{c}{\textbf{MV3DCD-ZS}~\cite{galappaththige2025multi}} & \multicolumn{3}{c}{\textbf{MV3DCD}~\cite{galappaththige2025multi}} & \multicolumn{3}{c}{\textbf{Online-SCD~\cite{galappaththige2025changes}}} \\
        \cmidrule(lr){2-4} \cmidrule(lr){5-7} \cmidrule(lr){8-10} \cmidrule(lr){11-13}
        \textbf{Metric} & Base & + Ours & $\Delta(\%)$ & Base & + Ours & $\Delta(\%)$ & Base & + Ours & $\Delta(\%)$ & Base & + Ours & $\Delta(\%)$ \\
        \midrule
        mIoU $\ouruparrow$ & 0.278 & \tabhl{0.359} & 29.1 & 0.382 & \tabhl{0.439} & 14.9 & 0.470 & \tabhl{0.498} & 6.0 & 0.486 & \tabhl{0.498} & 2.5 \\
        F1 $\ouruparrow$   & 0.402 & \tabhl{0.502} & 24.9 & 0.526 & \tabhl{0.593} & 12.7 & 0.621 & \tabhl{0.649} & 4.5 & 0.638 & \tabhl{0.651} & 2.1 \\
        \bottomrule
    \end{tabular}}
\end{table}

\begin{table}[t]
    \centering
    \caption{Quantitative results for pose-agnostic AD on MAD-Real~\cite{zhou2023pad}. Our UE guidance enables SplatPose~\cite{kruse2024splatpose} to match or surpass its successor, SplatPosePlus~\cite{liu2024splatpose+}. Furthermore, applying our framework directly to SplatPosePlus advances the current state-of-the-art. Percentage increase over the baseline is given in the $\Delta(\%)$ column.}
    \label{tab:anomaly_detection}
     \scalebox{0.85}{
    \setlength{\tabcolsep}{6pt}
    \begin{tabular}{@{}lcccccc@{}}
        \toprule
        & \multicolumn{3}{c}{\textbf{SplatPose}~\cite{kruse2024splatpose}} & \multicolumn{3}{c}{\textbf{SplatPosePlus}~\cite{liu2024splatpose+}} \\
        \cmidrule(lr){2-4} \cmidrule(lr){5-7}
        \textbf{Metric} & Base & + Ours & $\Delta(\%)$ & Base & + Ours & $\Delta(\%)$ \\
        \midrule
        AUROC $\ouruparrow$ & 0.929 & \tabhl{0.939} & 1.1 & 0.940 & \tabhl{0.956} & 0.7 \\
        AUPRO $\ouruparrow$ & 0.700 & \tabhl{0.765} & 9.3 & 0.761 & \tabhl{0.798} & 4.9 \\
        \bottomrule
    \end{tabular}}
\end{table}

\section{Limitations}
\label{sec:limitations}

Our approach has two main limitations. First, due to its strict post-hoc nature, the spatial granularity of our uncertainty maps is bottlenecked by the density and scale of the underlying Gaussians: in coarsely reconstructed regions (few large primitives), the predicted uncertainty is equally coarse. We do not alter primitives, as our goal is a lightweight estimator preserving model fidelity and structure. Second, we estimate a single \emph{predictive} uncertainty by treating residuals as a proxy for the expected photometric error; this conflates aleatoric, epistemic, and optimization-related effects rather than disentangling them, and does not yield a probabilistic distribution. These limitations are not unique; existing approaches also lack disentanglement, while only stochastic formulations~\cite{li2024variational,lyu2024manifold} provide a probabilistic form. We further study controlled aleatoric and epistemic perturbations in App.~G and find our UE behaves as expected.

\section{Conclusion}
\label{sec:conclusion}

We introduced a lightweight, plug-and-play framework for predictive photometric uncertainty estimation in 3DGS-based NVS. By formulating post-hoc uncertainty estimation as a linear least-squares problem with Bayesian-inspired regularization, our method effectively predicts pixel-level uncertainty maps under both dense and sparse captures. This enables state-of-the-art results in active view selection, pose-agnostic scene change detection, and anomaly detection. 

\section*{Acknowledgment}
This work was supported by the Australian Research Council Research Hub in Intelligent Robotic Systems for Real-Time Asset Management (IH210100030) (ARIAM) and Abyss Solutions. C.J., N.S., and D.M. also acknowledge ongoing support from the QUT Centre for Robotics.
T.G.\ P.S., and M.R.\ acknowledge support by the state of North Rhine-Westphalia and the European Union within the EFRE/JTF project ``Just scan it 3D'', grant no.\ EFRE-20800529.
E.H.\ and M.R.\ acknowledge support through the junior research group project ``UnrEAL'' by the German Federal Ministry of Education and Research (BMBF), grant no.\ 01IS22069.
M.R.\ also acknowledges mobility support by the German Academic Exchange Service (DAAD PPP), grant no.\ 57700453.

%
%
\bibliographystyle{splncs04}

\bibliography{main}

@String(CVPR= {IEEE Conf. Comput. Vis. Pattern Recog.})

@String(ICCV= {Int. Conf. Comput. Vis.})

@String(ECCV= {Eur. Conf. Comput. Vis.})

@String(BMVC= {Brit. Mach. Vis. Conf.})

@String(TOG= {ACM Trans. Graph.})

@String(ICASSP=	{ICASSP})

@String(ICIP = {IEEE Int. Conf. Image Process.})

@String(AAAI = {AAAI})

@String(CVPR  = {CVPR})

@String(ICCV  = {ICCV})

@String(ECCV  = {ECCV})

@String(BMVC  =	{BMVC})

@String(TOG   = {ACM TOG})

@String(ICIP  = {ICIP})

@inproceedings{zwicker2001ewa,
  title={EWA volume splatting},
  author={Zwicker, Matthias and Pfister, Hanspeter and Van Baar, Jeroen and Gross, Markus},
  booktitle={Proceedings Visualization, 2001. VIS'01.},
  pages={29--36},
  year={2001},
  organization={IEEE}
}

@article{kerbl20233dgaussiansplatting,
  title={3D Gaussian Splatting for Real-Time Radiance Field Rendering},
  author={Kerbl, Bernhard and Kopanas, Georgios and Leimkuehler, Thomas and Drettakis, George},
  journal={ACM Transactions on Graphics (TOG)},
  volume={42},
  number={4},
  pages={1--14},
  year={2023},
  publisher={ACM New York, NY, USA}
}

@article{mildenhall2021nerf,
  title={Nerf: Representing scenes as neural radiance fields for view synthesis},
  author={Mildenhall, Ben and Srinivasan, Pratul P and Tancik, Matthew and Barron, Jonathan T and Ramamoorthi, Ravi and Ng, Ren},
  journal={Communications of the ACM},
  volume={65},
  number={1},
  pages={99--106},
  year={2021},
  publisher={ACM New York, NY, USA}
}

@article{kheradmand20243d,
  title={3d gaussian splatting as markov chain monte carlo},
  author={Kheradmand, Shakiba and Rebain, Daniel and Sharma, Gopal and Sun, Weiwei and Tseng, Yang-Che and Isack, Hossam and Kar, Abhishek and Tagliasacchi, Andrea and Yi, Kwang Moo},
  journal={Advances in Neural Information Processing Systems},
  volume={37},
  pages={80965--80986},
  year={2024}
}

@inproceedings{ye2024absgs,
  title={Absgs: Recovering fine details in 3d gaussian splatting},
  author={Ye, Zongxin and Li, Wenyu and Liu, Sidun and Qiao, Peng and Dou, Yong},
  booktitle={Proceedings of the 32nd ACM International Conference on Multimedia},
  pages={1053--1061},
  year={2024}
}

@inproceedings{chung2024depth,
  title={Depth-regularized optimization for 3d gaussian splatting in few-shot images},
  author={Chung, Jaeyoung and Oh, Jeongtaek and Lee, Kyoung Mu},
  booktitle={Proceedings of the IEEE/CVF Conference on Computer Vision and Pattern Recognition},
  pages={811--820},
  year={2024}
}

@article{kerbl2024hierarchical,
  title={A hierarchical 3d gaussian representation for real-time rendering of very large datasets},
  author={Kerbl, Bernhard and Meuleman, Andreas and Kopanas, Georgios and Wimmer, Michael and Lanvin, Alexandre and Drettakis, George},
  journal={ACM Transactions on Graphics (TOG)},
  volume={43},
  number={4},
  pages={1--15},
  year={2024},
  publisher={ACM New York, NY, USA}
}

@inproceedings{hanson2025speedy,
  title={{Speedy-splat}: Fast 3d gaussian splatting with sparse pixels and sparse primitives},
  author={Hanson, Alex and Tu, Allen and Lin, Geng and Singla, Vasu and Zwicker, Matthias and Goldstein, Tom},
  booktitle={Proceedings of the Computer Vision and Pattern Recognition Conference},
  pages={21537--21546},
  year={2025}
}

@inproceedings{mallick2024taming,
  title={Taming 3dgs: High-quality radiance fields with limited resources},
  author={Mallick, Saswat Subhajyoti and Goel, Rahul and Kerbl, Bernhard and Steinberger, Markus and Carrasco, Francisco Vicente and De La Torre, Fernando},
  booktitle={SIGGRAPH Asia 2024 Conference Papers},
  pages={1--11},
  year={2024}
}

@book{tarantola2005inverse,
  title={Inverse problem theory and methods for model parameter estimation},
  author={Tarantola, Albert},
  year={2005},
  publisher={SIAM}
}

@inproceedings{lyu2024manifold,
  title={Manifold sampling for differentiable uncertainty in radiance fields},
  author={Lyu, Linjie and Tewari, Ayush and Habermann, Marc and Saito, Shunsuke and Zollh{\"o}fer, Michael and Leimk{\"u}hler, Thomas and Theobalt, Christian},
  booktitle={SIGGRAPH Asia 2024 Conference Papers},
  pages={1--11},
  year={2024}
}

@inproceedings{shen2021stochastic,
  title={Stochastic neural radiance fields: Quantifying uncertainty in implicit 3d representations},
  author={Shen, Jianxiong and Ruiz, Adria and Agudo, Antonio and Moreno-Noguer, Francesc},
  booktitle={2021 International Conference on 3D Vision (3DV)},
  pages={972--981},
  year={2021},
  organization={IEEE}
}

@inproceedings{jiang2024fisherrf,
  title={Fisherrf: Active view selection and mapping with radiance fields using fisher information},
  author={Jiang, Wen and Lei, Boshu and Daniilidis, Kostas},
  booktitle={European Conference on Computer Vision},
  pages={422--440},
  year={2024},
  organization={Springer}
}

@article{li2024variational,
  title={Variational multi-scale representation for estimating uncertainty in 3d gaussian splatting},
  author={Li, Ruiqi and Cheung, Yiu-ming},
  journal={Advances in Neural Information Processing Systems},
  volume={37},
  pages={87934--87958},
  year={2024}
}

@article{wang2004imagequality,
  title={Image quality assessment: from error visibility to structural similarity},
  author={Wang, Zhou and Bovik, Alan C and Sheikh, Hamid R and Simoncelli, Eero P},
  journal={IEEE transactions on image processing},
  volume={13},
  number={4},
  pages={600--612},
  year={2004},
  publisher={IEEE}
}

@article{hedman2018deepblending,
  author = {Hedman, Peter and Philip, Julien and Price, True and Frahm, Jan-Michael and Drettakis, George and Brostow, Gabriel},
  title = {Deep Blending for Free-viewpoint Image-based Rendering},
  journal = {ACM Transactions on Graphics (Proc. SIGGRAPH Asia)},
  publisher = {ACM},
  volume    = {37},
  number    = {6},
  pages     = {257:1--257:15},
  year      = {2018}
}

@article{knapitsch2017tanksandtemples,
    author    = {Arno Knapitsch and Jaesik Park and Qian-Yi Zhou and Vladlen Koltun},
    title     = {Tanks and Temples: Benchmarking Large-Scale Scene Reconstruction},
    journal   = {ACM Transactions on Graphics},
    volume    = {36},
    number    = {4},
    year      = {2017},
}

@inproceedings{ilg2018uncertainty,
  title={Uncertainty estimates and multi-hypotheses networks for optical flow},
  author={Ilg, Eddy and Cicek, Ozgun and Galesso, Silvio and Klein, Aaron and Makansi, Osama and Hutter, Frank and Brox, Thomas},
  booktitle={Proceedings of the European Conference on Computer Vision (ECCV)},
  pages={652--667},
  year={2018}
}

@inproceedings{shen2022conditional,
  title={Conditional-flow nerf: Accurate 3d modelling with reliable uncertainty quantification},
  author={Shen, Jianxiong and Agudo, Antonio and Moreno-Noguer, Francesc and Ruiz, Adria},
  booktitle={European Conference on Computer Vision},
  pages={540--557},
  year={2022},
  organization={Springer}
}

@INPROCEEDINGS{suenderhauf2023densityaware,
  author={Sünderhauf, Niko and Abou-Chakra, Jad and Miller, Dimity},
  booktitle={2023 IEEE International Conference on Robotics and Automation (ICRA)}, 
  title={Density-aware NeRF Ensembles: Quantifying Predictive Uncertainty in Neural Radiance Fields}, 
  year={2023},
  volume={},
  number={},
  pages={9370-9376}
}

@InProceedings{seo2023flipnerf,
    author    = {Seo, Seunghyeon and Chang, Yeonjin and Kwak, Nojun},
    title     = {FlipNeRF: Flipped Reflection Rays for Few-shot Novel View Synthesis},
    booktitle = {Proceedings of the IEEE/CVF International Conference on Computer Vision (ICCV)},
    year      = {2023},
    pages     = {22883-22893}
}

@inproceedings{goli2024bayesrays,
  title={Bayes' rays: Uncertainty quantification for neural radiance fields},
  author={Goli, Lily and Reading, Cody and Sell{\'a}n, Silvia and Jacobson, Alec and Tagliasacchi, Andrea},
  booktitle={Proceedings of the IEEE/CVF Conference on Computer Vision and Pattern Recognition},
  pages={20061--20070},
  year={2024}
}

@article{nakayama2024provnerf,
  title={ProvNeRF: Modeling per Point Provenance in NeRFs as a Stochastic Field},
  author={Nakayama, Kiyohiro and Uy, Mikaela Angelina and You, Yang and Li, Ke and Guibas, Leonidas J},
  journal={Advances in Neural Information Processing Systems},
  volume={37},
  pages={99145--99160},
  year={2024}
}

@INPROCEEDINGS{wilson2024modelinguncertainty3dgaussian,
  author={Wilson, Joey and Almeida, Marcelino and Sun, Min and Mahajan, Sachit and Ghaffari, Maani and Ewen, Parker and Ghasemalizadeh, Omid and Kuo, Cheng-Hao and Sen, Arnie},
  booktitle={2025 IEEE International Conference on Robotics and Automation (ICRA)}, 
  title={Modeling Uncertainty in 3D Gaussian Splatting Through Continuous Semantic Splatting}, 
  year={2025},
  volume={},
  number={},
  pages={3284-3290}}

@inproceedings{rota2025revising,
author = {Rota Bul\`{o}, Samuel and Porzi, Lorenzo and Kontschieder, Peter},
title = {Revising Densification in Gaussian Splatting},
year = {2024},
booktitle = {Proceedings of the European conference on computer vision (ECCV)},
pages = {347–362},
numpages = {16}
}

@inproceedings{wilson2025popgs,
  title = {{{POp-GS}}: {{Next Best View}} in {{3D-Gaussian Splatting}} with {{P-Optimality}}},
  shorttitle = {{{POp-GS}}},
  booktitle = {Proceedings of the Computer Vision and Pattern Recognition Conference (CVPR)},
  author = {Wilson, Joey and Almeida, Marcelino and Mahajan, Sachit and Labrie, Martin and Ghaffari, Maani and Ghasemalizadeh, Omid and Sun, Min and Kuo, Cheng-Hao and Sen, Arnab},
  year = 2025,
  pages = {3646--3655}
}

@ARTICLE{aria2025modelinguncertainty,
  author={Aira, Luca Savant and Valsesia, Diego and Magli, Enrico},
  journal={IEEE Transactions on Neural Networks and Learning Systems}, 
  title={Modeling Uncertainty for Gaussian Splatting}, 
  year={2025},
  volume={36},
  number={6},
  pages={11657-11663}
}

@InProceedings{pan2022activenerf,
author="Pan, Xuran
and Lai, Zihang
and Song, Shiji
and Huang, Gao",
editor="Avidan, Shai
and Brostow, Gabriel
and Ciss{\'e}, Moustapha
and Farinella, Giovanni Maria
and Hassner, Tal",
title="ActiveNeRF: Learning Where to See with Uncertainty Estimation",
booktitle="Computer Vision -- ECCV 2022",
year="2022",
publisher="Springer Nature Switzerland",
address="Cham",
pages="230--246",
isbn="978-3-031-19827-4"
}

@inproceedings{kopanas2023improvingnerf,
booktitle = {Vision, Modeling, and Visualization},
editor = {Guthe, Michael and Grosch, Thorsten},
title = {{Improving NeRF Quality by Progressive Camera Placement for Free-Viewpoint Navigation}},
author = {Kopanas, Georgios and Drettakis, George},
year = {2023},
publisher = {The Eurographics Association},
ISBN = {978-3-03868-232-5}
}

@InProceedings{barron2022mipnerf,
    author    = {Barron, Jonathan T. and Mildenhall, Ben and Verbin, Dor and Srinivasan, Pratul P. and Hedman, Peter},
    title     = {Mip-NeRF 360: Unbounded Anti-Aliased Neural Radiance Fields},
    booktitle = {Proceedings of the IEEE/CVF Conference on Computer Vision and Pattern Recognition (CVPR)},
    year      = {2022},
    pages     = {5470-5479}
}

@inproceedings{matsuki2024gaussian,
  title={Gaussian splatting slam},
  author={Matsuki, Hidenobu and Murai, Riku and Kelly, Paul HJ and Davison, Andrew J},
  booktitle={Proceedings of the IEEE/CVF Conference on Computer Vision and Pattern Recognition},
  pages={18039--18048},
  year={2024}
}

@inproceedings{galappaththige2025multi,
  title={Multi-View Pose-Agnostic Change Localization with Zero Labels},
  author={Galappaththige, Chamuditha Jayanga and Lai, Jason and Windrim, Lloyd and Dansereau, Donald and Sunderhauf, Niko and Miller, Dimity},
  booktitle={Proceedings of the Computer Vision and Pattern Recognition Conference},
  pages={11600--11610},
  year={2025}
}

@article{alcantarilla2018street,
  title={Street-view change detection with deconvolutional networks},
  author={Alcantarilla, Pablo F and Stent, Simon and Ros, German and Arroyo, Roberto and Gherardi, Riccardo},
  journal={Autonomous Robots},
  volume={42},
  number={7},
  pages={1301--1322},
  year={2018},
  publisher={Springer}
}

@inproceedings{daudt2018fully,
  title={Fully convolutional siamese networks for change detection},
  author={Daudt, Rodrigo Caye and Le Saux, Bertr and Boulch, Alexandre},
  booktitle={2018 25th IEEE international conference on image processing (ICIP)},
  pages={4063--4067},
  year={2018},
  organization={IEEE}
}

@inproceedings{sakurada2015change,
  title={Change Detection from a Street Image Pair using CNN Features and Superpixel Segmentation},
  author={Sakurada, Ken and Okatani, Takayuki},
  booktitle={Proceedings of the British Machine Vision Conference (BMVC)},
  year={2015},
  pages = {61.1-61.12}
}

@inproceedings{lin2025robust,
  title={Robust scene change detection using visual foundation models and cross-attention mechanisms},
  author={Lin, Chun-Jung and Garg, Sourav and Chin, Tat-Jun and Dayoub, Feras},
  booktitle={2025 IEEE International Conference on Robotics and Automation (ICRA)},
  pages={8337--8343},
  year={2025},
  organization={IEEE}
}

@inproceedings{cyws2d,
  title={The change you want to see},
  author={Sachdeva, Ragav and Zisserman, Andrew},
  booktitle={Proceedings of the IEEE/CVF Winter Conference on Applications of Computer Vision},
  pages={3993--4002},
  year={2023}
}

@inproceedings{cyws3d,
  title={The change you want to see (now in 3d)},
  author={Sachdeva, Ragav and Zisserman, Andrew},
  booktitle={Proceedings of the IEEE/CVF International Conference on Computer Vision Workshops},
  pages={2060--2069},
  year={2023}
}

@inproceedings{kim2025towards,
  title={Towards Generalizable Scene Change Detection},
  author={Kim, Jae-Woo and Kim, Ue-Hwan},
  booktitle={Proceedings of the Computer Vision and Pattern Recognition Conference},
  pages={24463--24473},
  year={2025}
}

@inproceedings{cho2025zero,
  title={Zero-shot scene change detection},
  author={Cho, Kyusik and Kim, Dong Yeop and Kim, Euntai},
  booktitle={Proceedings of the AAAI Conference on Artificial Intelligence},
  volume={39 (3)},
  pages={2509--2517},
  year={2025}
}

@article{lu20253dgs,
  title={3dgs-cd: 3d gaussian splatting-based change detection for physical object rearrangement},
  author={Lu, Ziqi and Ye, Jianbo and Leonard, John},
  journal={IEEE Robotics and Automation Letters},
  year={2025},
  publisher={IEEE}
}

@inproceedings{jiang2025gaussian,
  title={Gaussian Difference: Find Any Change Instance in 3D Scenes},
  author={Jiang, Binbin and Huang, Rui and Zhao, Qingyi and Zhang, Yuxiang},
  booktitle={ICASSP 2025-2025 IEEE International Conference on Acoustics, Speech and Signal Processing (ICASSP)},
  pages={1--5},
  year={2025},
  organization={IEEE}
}

@article{zhou2023pad,
  title={Pad: A dataset and benchmark for pose-agnostic anomaly detection},
  author={Zhou, Qiang and Li, Weize and Jiang, Lihan and Wang, Guoliang and Zhou, Guyue and Zhang, Shanghang and Zhao, Hao},
  journal={Advances in Neural Information Processing Systems},
  volume={36},
  pages={44558--44571},
  year={2023}
}

@inproceedings{schonberger2016structure,
  title={Structure-from-motion revisited},
  author={Schonberger, Johannes L and Frahm, Jan-Michael},
  booktitle={Proceedings of the IEEE conference on computer vision and pattern recognition},
  pages={4104--4113},
  year={2016}
}

@inproceedings{kruse2024splatpose,
  title={Splatpose \& detect: Pose-agnostic 3d anomaly detection},
  author={Kruse, Mathis and Rudolph, Marco and Woiwode, Dominik and Rosenhahn, Bodo},
  booktitle={Proceedings of the IEEE/CVF Conference on Computer Vision and Pattern Recognition},
  pages={3950--3960},
  year={2024}
}

@inproceedings{kirillov2023segment,
  title={Segment anything},
  author={Kirillov, Alexander and Mintun, Eric and Ravi, Nikhila and Mao, Hanzi and Rolland, Chloe and Gustafson, Laura and Xiao, Tete and Whitehead, Spencer and Berg, Alexander C and Lo, Wan-Yen and others},
  booktitle={Proceedings of the IEEE/CVF international conference on computer vision},
  pages={4015--4026},
  year={2023}
}

@article{oquab2023dinov2,
  title={Dinov2: Learning robust visual features without supervision},
  author={Oquab, Maxime and Darcet, Timoth{\'e}e and Moutakanni, Th{\'e}o and Vo, Huy and Szafraniec, Marc and Khalidov, Vasil and Fernandez, Pierre and Haziza, Daniel and Massa, Francisco and El-Nouby, Alaaeldin and others},
  journal={arXiv preprint arXiv:2304.07193},
  year={2023}
}

@INPROCEEDINGS{kannanZero,
  author={Kannan, Shyam Sundar and Min, Byung-Cheol},
  booktitle={2025 IEEE International Conference on Robotics and Automation (ICRA)}, 
  title={ZeroSCD: Zero-Shot Street Scene Change Detection}, 
  year={2025},
  volume={},
  number={},
  pages={4665-4671}
}

@inproceedings{chen2021dr,
  title={Dr-tanet: Dynamic receptive temporal attention network for street scene change detection},
  author={Chen, Shuo and Yang, Kailun and Stiefelhagen, Rainer},
  booktitle={2021 IEEE Intelligent Vehicles Symposium (IV)},
  pages={502--509},
  year={2021},
  organization={IEEE}
}

@inproceedings{varghese2018changenet,
  title={ChangeNet: A deep learning architecture for visual change detection},
  author={Varghese, Ashley and Gubbi, Jayavardhana and Ramaswamy, Akshaya and Balamuralidhar, P},
  booktitle={Proceedings of the European conference on computer vision (ECCV) workshops},
  pages={0--0},
  year={2018}
}

@article{lei2020hierarchical,
  title={Hierarchical paired channel fusion network for street scene change detection},
  author={Lei, Yinjie and Peng, Duo and Zhang, Pingping and Ke, Qiuhong and Li, Haifeng},
  journal={IEEE Transactions on Image Processing},
  volume={30},
  pages={55--67},
  year={2020},
  publisher={IEEE}
}

@inproceedings{liu2024splatpose+,
  title={SplatPose+: Real-Time Image-Based Pose-Agnostic 3D Anomaly Detection},
  author={Liu, Yizhe and Hu, Yan Song and Chen, Yuhao and Zelek, John},
  booktitle={European Conference on Computer Vision Workshops},
  pages={378--391},
  year={2024},
  organization={Springer}
}

@inproceedings{wu20244d,
  title={4d gaussian splatting for real-time dynamic scene rendering},
  author={Wu, Guanjun and Yi, Taoran and Fang, Jiemin and Xie, Lingxi and Zhang, Xiaopeng and Wei, Wei and Liu, Wenyu and Tian, Qi and Wang, Xinggang},
  booktitle={Proceedings of the IEEE/CVF conference on computer vision and pattern recognition},
  pages={20310--20320},
  year={2024}
}

@InProceedings{hanson2025pup3dgs,
    author    = {Hanson, Alex and Tu, Allen and Singla, Vasu and Jayawardhana, Mayuka and Zwicker, Matthias and Goldstein, Tom},
    title     = {{PUP 3D-GS}: Principled Uncertainty Pruning for 3D Gaussian Splatting},
    booktitle = {Proceedings of the Computer Vision and Pattern Recognition Conference (CVPR)},
    year      = {2025},
    pages     = {5949-5958}
}

@inproceedings{alpherts2025emplace,
  title={Emplace: Self-supervised urban scene change detection},
  author={Alpherts, Tim and Ghebreab, Sennay and van Noord, Nanne},
  booktitle={Proceedings of the AAAI Conference on Artificial Intelligence},
  volume={39 (2)},
  pages={1737--1745},
  year={2025}
}

@article{kowsky1986quadrature,
  title={A quadrature formula over the sphere with application to high resolution spherical harmonic analysis},
  author={Kowsky, WS},
  journal={Bull. Gdod},
  volume={60},
  pages={1--14},
  year={1986},
  publisher={Springer}
}

@book{bishop2006pattern,
  title={Pattern recognition and machine learning},
  author={Bishop, Christopher M and Nasrabadi, Nasser M},
  volume={4},
  year={2006},
  publisher={Springer}
}

@inproceedings{gottwald2025primu,
  title={PRIMU: Uncertainty Estimation for Novel Views in Gaussian Splatting from Primitive-Based Representations of Error and Coverage},
  author={Gottwald, Thomas and Heinert, Edgar and Stehr, Peter and Galappaththige, Chamuditha Jayanga and Rottmann, Matthias},
  booktitle={Proceedings of the IEEE/CVF Conference on Computer Vision and Pattern Recognition},
  pages={11871--11880},
  year={2026}
}

@inproceedings{zhu2025loopsplat,
  title={Loopsplat: Loop closure by registering 3d gaussian splats},
  author={Zhu, Liyuan and Li, Yue and Sandstr{\"o}m, Erik and Huang, Shengyu and Schindler, Konrad and Armeni, Iro},
  booktitle={2025 International Conference on 3D Vision (3DV)},
  pages={156--167},
  year={2025},
  organization={IEEE}
}

@article{fei20243d,
  title={3d gaussian splatting as new era: A survey},
  author={Fei, Ben and Xu, Jingyi and Zhang, Rui and Zhou, Qingyuan and Yang, Weidong and He, Ying},
  journal={IEEE Transactions on Visualization and Computer Graphics},
  year={2024},
  publisher={IEEE}
}

@article{Bergmann2019MVTecA,
  title={MVTec AD — A Comprehensive Real-World Dataset for Unsupervised Anomaly Detection},
  author={Paul Bergmann and Michael Fauser and David Sattlegger and Carsten Steger},
  journal={2019 IEEE/CVF Conference on Computer Vision and Pattern Recognition (CVPR)},
  year={2019},
  pages={9584-9592}
}

@article{Roth2021TowardsTR,
  title={Towards Total Recall in Industrial Anomaly Detection},
  author={Karsten Roth and Latha Pemula and Joaquin Zepeda and Bernhard Scholkopf and Thomas Brox and Peter Gehler},
  journal={2022 IEEE/CVF Conference on Computer Vision and Pattern Recognition (CVPR)},
  year={2021},
  pages={14298-14308}
}

@inproceedings{galappaththige2025changes,
  title={Changes in Real Time: Online Scene Change Detection with Multi-View Fusion},
  author={Galappaththige, Chamuditha Jayanga and Lai, Jason and Windrim, Lloyd and Dansereau, Donald and Suenderhauf, Niko and Miller, Dimity},
  booktitle={Proceedings of the IEEE/CVF Conference on Computer Vision and Pattern Recognition},
  pages={32246--32256},
  year={2026}
}

@article{sunderhauf2018limits,
  title={The limits and potentials of deep learning for robotics},
  author={S{\"u}nderhauf, Niko and Brock, Oliver and Scheirer, Walter and Hadsell, Raia and Fox, Dieter and Leitner, J{\"u}rgen and Upcroft, Ben and Abbeel, Pieter and Burgard, Wolfram and Milford, Michael and others},
  journal={The International journal of robotics research},
  volume={37},
  number={4-5},
  pages={405--420},
  year={2018},
  publisher={SAGE Publications Sage UK: London, England}
}

\clearpage
\setcounter{page}{1}
\begin{center}
    \Large \textbf{
    Appendix -- Predictive Photometric Uncertainty in Gaussian Splatting for Novel View Synthesis
    }
\end{center}
\newcounter{mysection}
\renewcommand\thesection{\setcounter{mysection}{\value{section}}\addtocounter{mysection}{-5}\Alph{mysection}}

\noindent This appendix provides supplementary technical and experimental details supporting the main findings of our paper.
In Appendix~\ref{sec:sup:viewdependentformlation}, we present the explicit mathematical formulation for incorporating directional dependency into our UE method through the use of SH and show that this does not affect the linearity of our least-squares formulation.
Appendix~\ref{sec:sup:3dgsvariants} details an ablation study evaluating our UE technique across various 3DGS variants.
Appendix~\ref{sec:sup:sparseViewUE} provides scene-wise results for our study on uncertainty estimation in highly novel views.
We compare the reconstruction fidelity of stochastic 3DGS methods, which the baselines in our main UE study employ, and standard 3DGS in Appendix~\ref{sec:sup:splatQuality}.
Appendix~\ref{sec:sup:avs_training_iterations} investigates the effect of the training budget (number of iterations) for uncertainty channels prior to selecting the next best view in AVS.
Additional qualitative examples for UE are presented in Appendix~\ref{sec:sup:qualitativeexamples}.
Finally, in Appendix~\ref{sec:sup:uncertainty_response}, we validate that our predictive uncertainty responds as expected to controlled aleatoric and epistemic perturbations of the training views.

\section{View Dependent Per-primitive Uncertainty}\label{sec:sup:viewdependentformlation}
We show that incorporating SH to model the directional dependence of uncertainty does not compromise the linearity of the least-squares problem described in \cref*{sec:un}.
To this end, we replace the per-Gaussian scalar uncertainties by SH coefficients and expand the uncertainty vector \(u\) accordingly, adjusting the matrix \(A\) to match this representation.

When using SH degree \(L\), there are \(s := (L+1)^2\) SH basis functions, typically denoted \(Y_{lm} : S^2 \to \mathbb{R}\) for \(l = 0, \dotsc, L\) and \(m = -l, \dotsc, l\). For a Gaussian \(G_k\), the direction-dependent uncertainty \(u_k : S^2 \to \mathbb{R}\) is represented using SH coefficients \(u_k^{lm}\) as
\begin{equation}\label{eq:sh_def}
    u_k(d) = \sum_{l=0}^L \sum_{m=-l}^l u_k^{lm} Y_{lm}(d).
\end{equation}
For notational convenience, we enumerate the basis functions as \(Y_1, \dotsc, Y_s\) and denote the corresponding coefficients of the \(k\)-th Gaussian as \(u_k^1, \dotsc, u_k^s\). This simplifies \cref{eq:sh_def} to
\begin{equation}\label{eq:sh_def_simplified}
    u_k(d) = \sum_{i=1}^s u_k^i Y_i(d).
\end{equation}
Define the uncertainty vector \(\Tilde{u} \in \mathbb{R}^{s \cdot K}\) by \(\Tilde{u}_{s (k-1) + i} = u_k^i\)
and the adjusted matrix \(A \in \mathbb{R}^{M \times s \cdot K}\) by
\begin{equation}
    A_{j,\, s (k-1) + i} = Y_i(d(x_j, G_k)) \, \alpha_k(x_j) \, \hat{T}_k(x_j)
\end{equation}
for \(j = 1, \dotsc, M\), \(k = 1, \dotsc, K\), and \(i = 1, \dotsc, s\), where \(d(x_j, G_k)\) denotes the viewing direction from pixel \(x_j\) toward \(G_k\). 
Using \cref{eq:sh_def_simplified}, we obtain
\begin{align}
    (A\Tilde{u})_j &= \sum_{k=1}^K \sum_{i=1}^s u_k^i Y_i(d(x_j, G_k)) \alpha_k(x_j) \hat{T}_k(x_j) \\
           &= \sum_{k=1}^K u_k(d(x_j, G_k)) \alpha_k(x_j) \hat{T}_k(x_j),
\end{align}
which is the rendered uncertainty at pixel \(x_j\).
Thus, estimating the SH coefficients that best explain the reconstruction residual \(y\) reduces to solving the linear least-squares problem
\begin{equation}
    \arg\min_u \| y - A \Tilde{u} \|_2^2,
\end{equation}
as in \cref*{eq:least_squares}.
Overall, the reason why directional dependence does not break linearity becomes evident: since the positions and covariances of the Gaussians remain fixed, the direction \(d(x_j, G_k)\) and thus its evaluation in the SH basis function $Y_i$ stays fixed. This allows us to incorporate these terms directly into the matrix $A$.

\begin{table}[t]
    \centering
    \caption{Quantitative results for UE in NVS on Mip-NeRF360~\citeNHL{barron2022mipnerf} with different 3DGS variants~\citeNHL{kerbl2024hierarchical,kerbl20233dgaussiansplatting,hanson2025speedy,ye2024absgs}. 
    Our approach can be applied post-hoc to any 3DGS variant and delivers comparable performance across variants.}
    \label{tab:ue_variants}
    \setlength{\tabcolsep}{3pt}
    \scalebox{0.75}{
    \begin{tabular}{lrrrr}
        \toprule
        \multirow{2}{*}{\textbf{Method}} &
        \multicolumn{2}{c}{AUSE$\ourdownarrow$} &
        \multicolumn{2}{c}{Pearson$\ouruparrow$} \\
        \cmidrule(lr){2-3} \cmidrule(lr){4-5}
        & L1 & DSSIM & L1 & DSSIM \\
        \midrule
        3DGS~\citeNHL{kerbl20233dgaussiansplatting}   &  &  &  &  \\
        $+$ FisherRF & 0.708 & 0.606 & -0.055 & 0.009  \\
        $+$ Ours & \tabhl{0.328} & \tabhl{0.214} & \tabhl{0.369} & \tabhl{0.547} \\
        \midrule
        Depth regularized 3DGS~\citeNHL{kerbl2024hierarchical}   &  &  &  &  \\
        $+$ FisherRF & 0.704 & 0.603 & -0.059 & -0.018 \\
        $+$ Ours & \tabhl{0.296} & \tabhl{0.170} & \tabhl{0.3805} & \tabhl{0.608} \\
        \midrule
        AbsGS~\citeNHL{ye2024absgs}      &   &  & & \\
        $+$ FisherRF & 0.871 & 0.819 &-0.014& -0.015 \\
        $+$ Ours & \tabhl{0.309} & \tabhl{0.226} & \tabhl{0.387} & \tabhl{0.507} \\
        \midrule
        Speedy-Splat~\citeNHL{hanson2025speedy} &  &  &  &  \\
        $+$ FisherRF & 0.766 & 0.591 & -0.009 & -0.003 \\
        $+$ Ours & \tabhl{0.297} & \tabhl{0.141} & \tabhl{0.368} & \tabhl{0.656} \\
        \bottomrule
    \end{tabular}}
\end{table}

\begin{table}[t]
    \centering
    \caption{Reconstruction quality comparison of stochastic 3DGS methods Manifold~\citeNHL{lyu2024manifold} and Var3DGS~\citeNHL{li2024variational} to standard 3DGS~\citeNHL{kerbl20233dgaussiansplatting}, which forms the underlying representation for our method and FisherRF~\citeNHL{jiang2024fisherrf}.
    Corresponding visual examples are shown in \cref{fig:sup:UEqualitative1,fig:sup:UEqualitative2}.
    }
    \label{tab:sup:quality}
    \setlength{\tabcolsep}{3pt}
    \scalebox{0.75}{
    \begin{tabular}{lrrrrrrrrr}
        \toprule
         \multirow{2}{*}{\textbf{Method}}  & \multicolumn{3}{c}{\textbf{Mip-NeRF360}} & \multicolumn{3}{c}{\textbf{Tanks\&Temples}} & \multicolumn{3}{c}{\textbf{Deep Blending}} \\
         \cmidrule(lr){2-4}\cmidrule(lr){5-7}\cmidrule(lr){8-10}
         & PSNR\(\ouruparrow\) & SSIM\(\ouruparrow\) & LPIPS\(\ourdownarrow\) & PSNR\(\ouruparrow\) & SSIM\(\ouruparrow\) & LPIPS\(\ourdownarrow\) & PSNR\(\ouruparrow\) & SSIM\(\ouruparrow\) & LPIPS\(\ourdownarrow\) \\
        \midrule
        Manifold~\citeNHL{lyu2024manifold} & 27.273 & 0.809 & 0.230 & \tabhl{23.517} & 0.842 & 0.190 & 29.239 & 0.902 & 0.252 \\
        Var3DGS~\citeNHL{li2024variational} & 27.137 & 0.807 & 0.231 & 23.321 & 0.840 & 0.186 & 28.806 & 0.895 & 0.263 \\
        3DGS~\citeNHL{kerbl20233dgaussiansplatting} & \tabhl{27.595} & \tabhl{0.816} & \tabhl{0.216} & 23.474 & \tabhl{0.848} & \tabhl{0.172} & \tabhl{29.899} & \tabhl{0.908} & \tabhl{0.242} \\
        \bottomrule
    \end{tabular}}
\end{table}

\section{UE Results on 3DGS Variants}\label{sec:sup:3dgsvariants}

Our method is fully plug-and-play and can be integrated with any 3DGS variant without architectural or optimization modifications. 
In \cref{tab:ue_variants}, we present a quantitative comparison across three representative extensions of 3DGS. 
First, we evaluate on depth-regularized 3DGS~\citeNHL{kerbl2024hierarchical}, which introduces geometric regularization to improve structural consistency. 
Second, we consider AbsGS~\citeNHL{ye2024absgs}, which employs novel adaptive density control to enhance rendering fidelity. 
Finally, we evaluate on Speedy-Splat~\citeNHL{hanson2025speedy}, a sparse-primitive formulation designed for faster training. 
Across all backbones, our method achieves UE performance comparable to or better than its performance on standard 3DGS, confirming its robustness and broad applicability.

\section{UE on Highly Novel Views: Scene-wise Results}\label{sec:sup:sparseViewUE}

\begin{figure}[t]
    \centering
    \includegraphics[width=\linewidth, trim=0.8cm 0.9cm 1.1cm 1.5cm, clip]{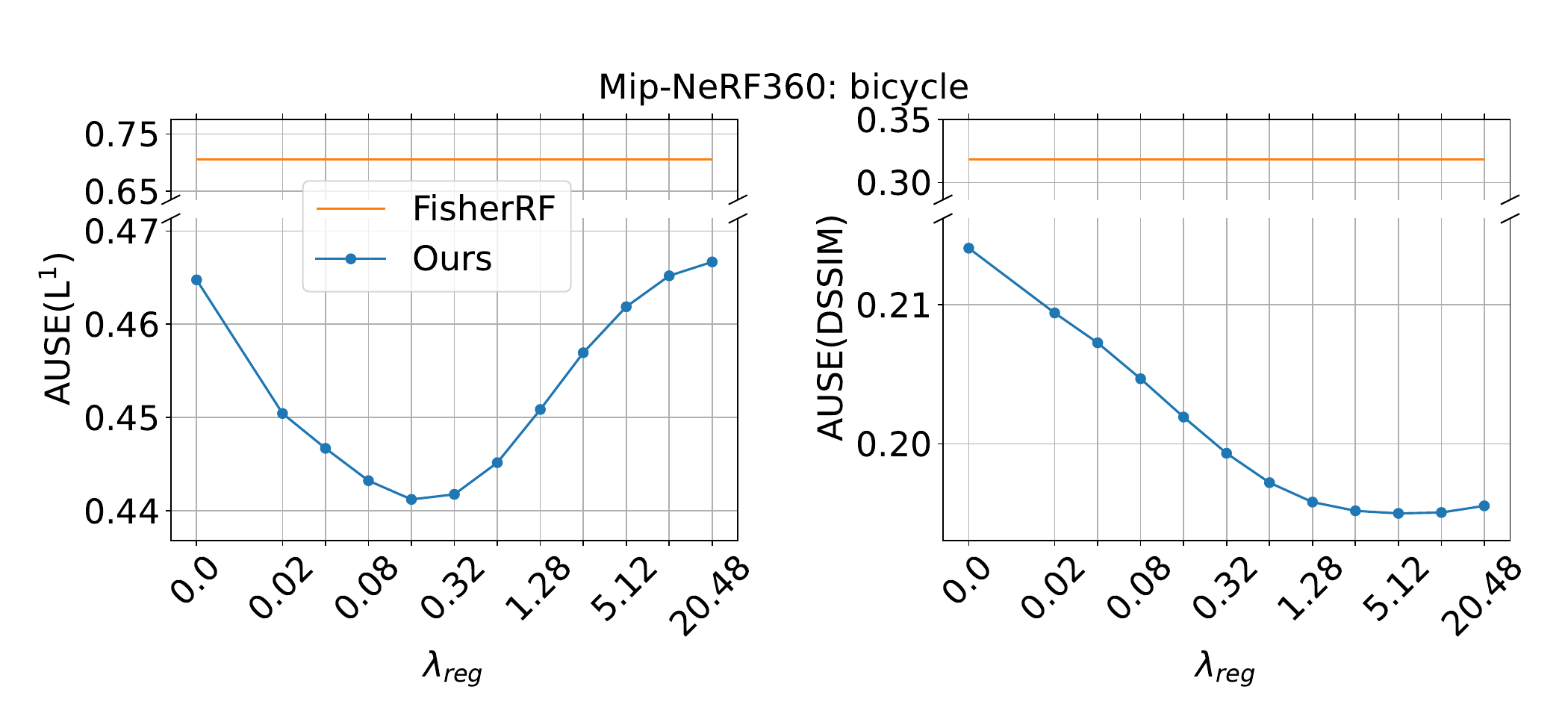}
    \includegraphics[width=\linewidth, trim=0.8cm 0.9cm 1.1cm 1.5cm, clip]{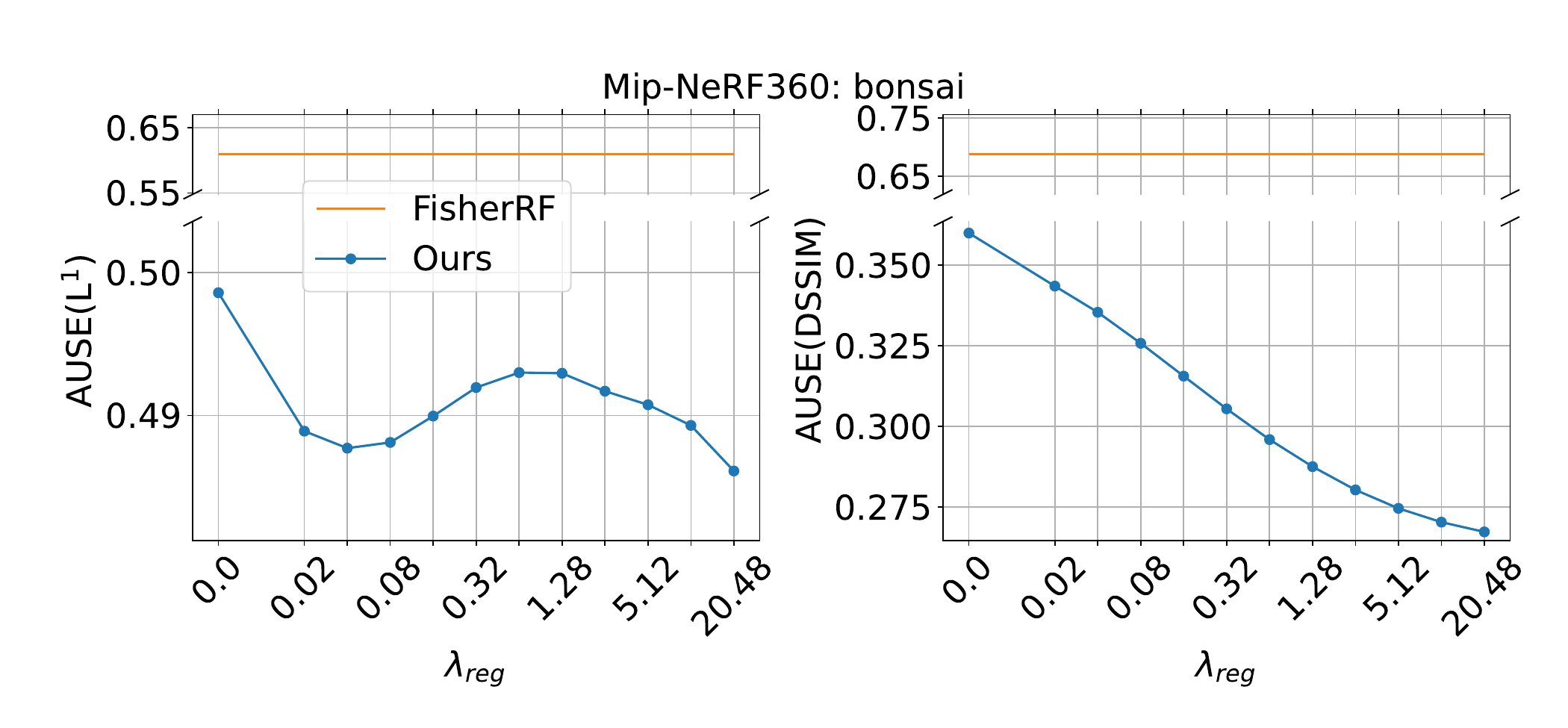}
    \includegraphics[width=\linewidth, trim=0.8cm 0.9cm 1.1cm 1.5cm, clip]{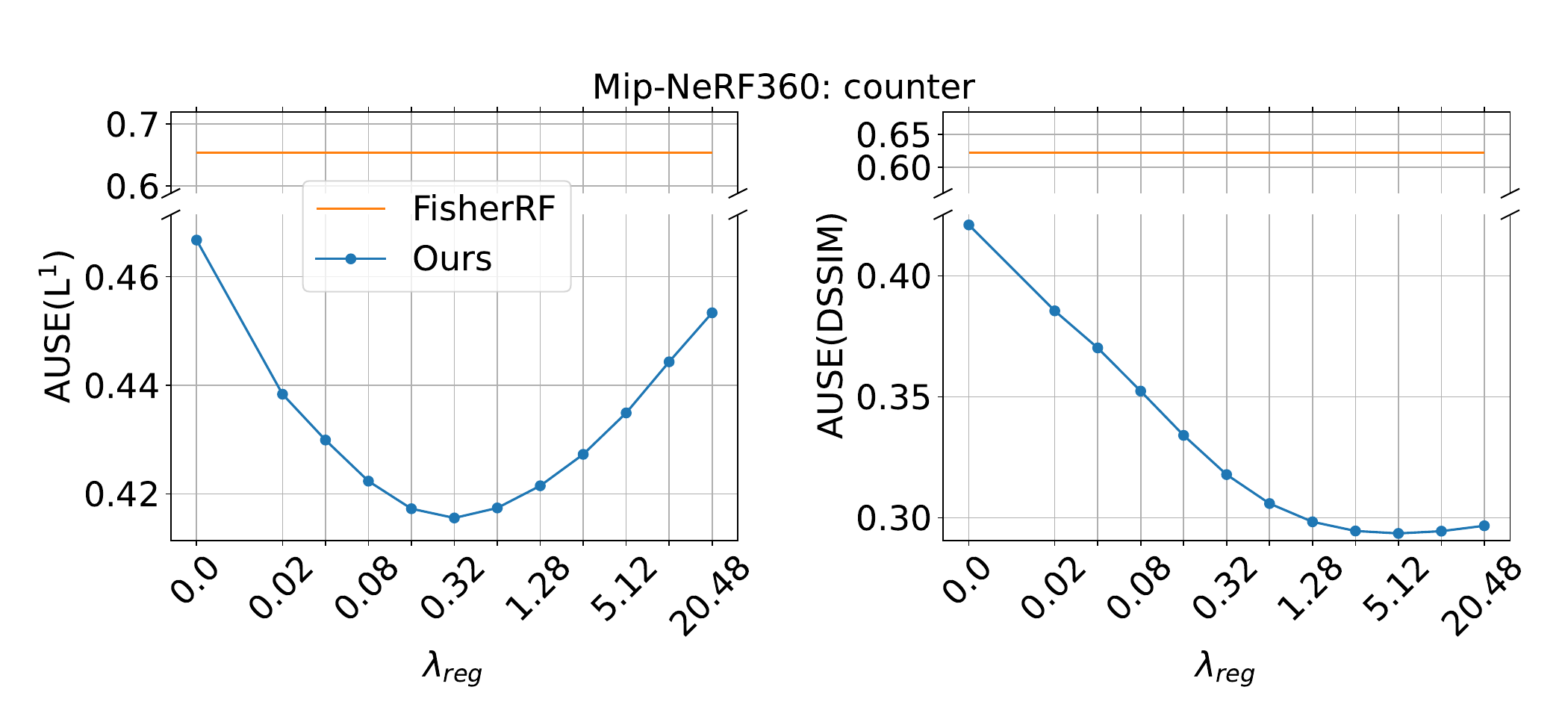}
    \caption{Scene-wise results for uncertainty estimation on highly novel views for the Mip-NeRF360~\citeNHL{barron2022mipnerf} scenes \textit{bicycle}, \textit{bonsai}, and \textit{counter}.}
    \label{fig:sup:sparseUEscenesPlot1}
\end{figure}

\begin{figure}[t]
    \centering
    \includegraphics[width=\linewidth, trim=0.8cm 0.9cm 1.1cm 1.5cm, clip]{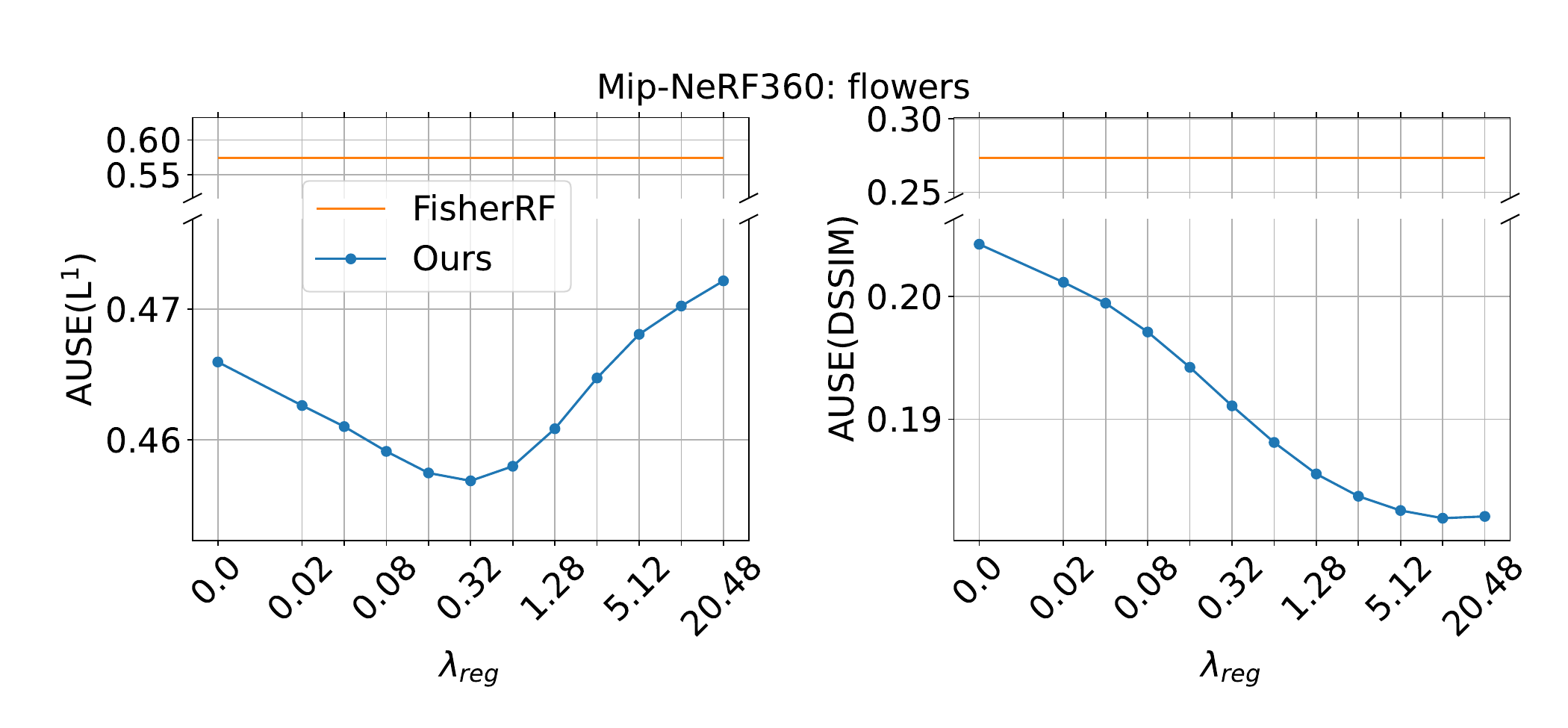}
    \includegraphics[width=\linewidth, trim=0.8cm 0.9cm 1.1cm 1.5cm, clip]{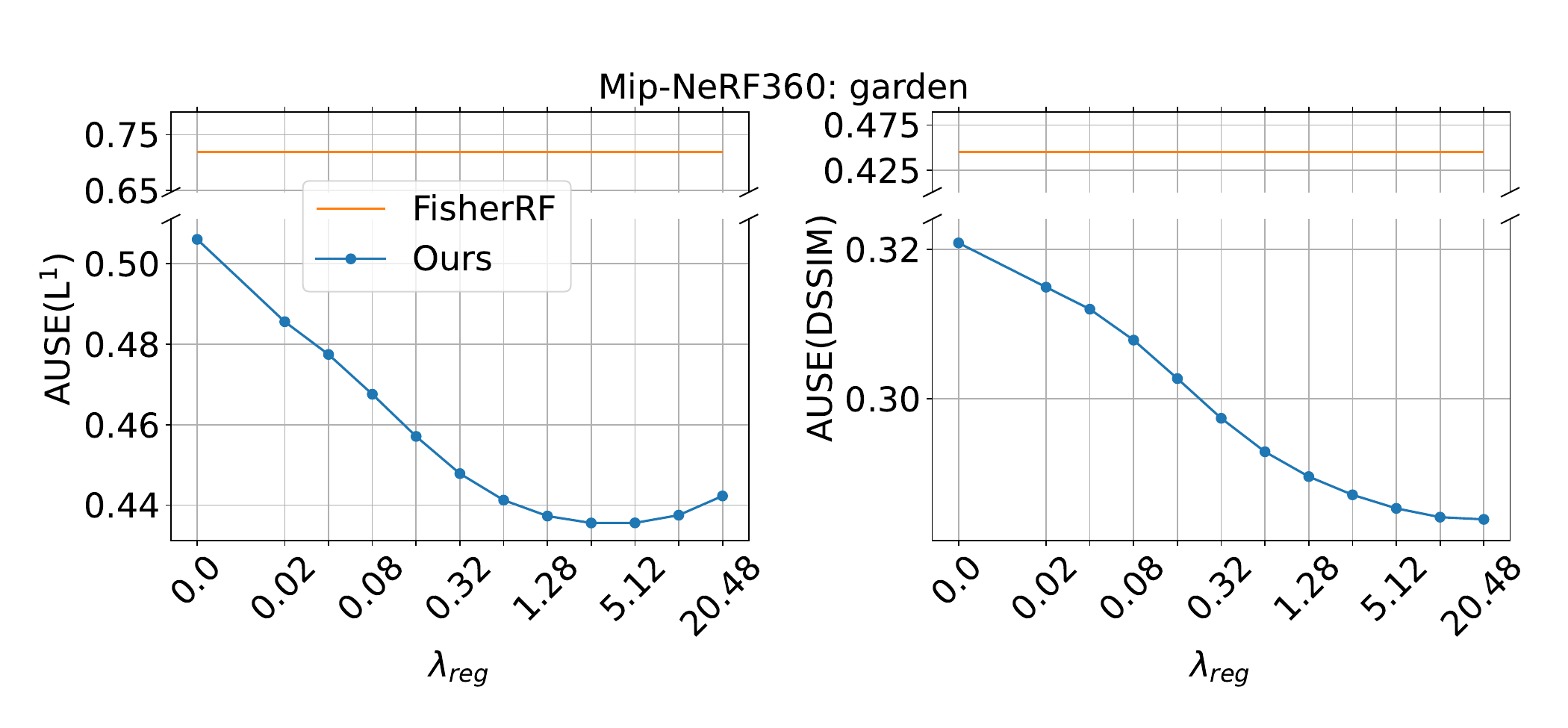}
    \includegraphics[width=\linewidth, trim=0.4cm 0.9cm 1.1cm 1.5cm, clip]{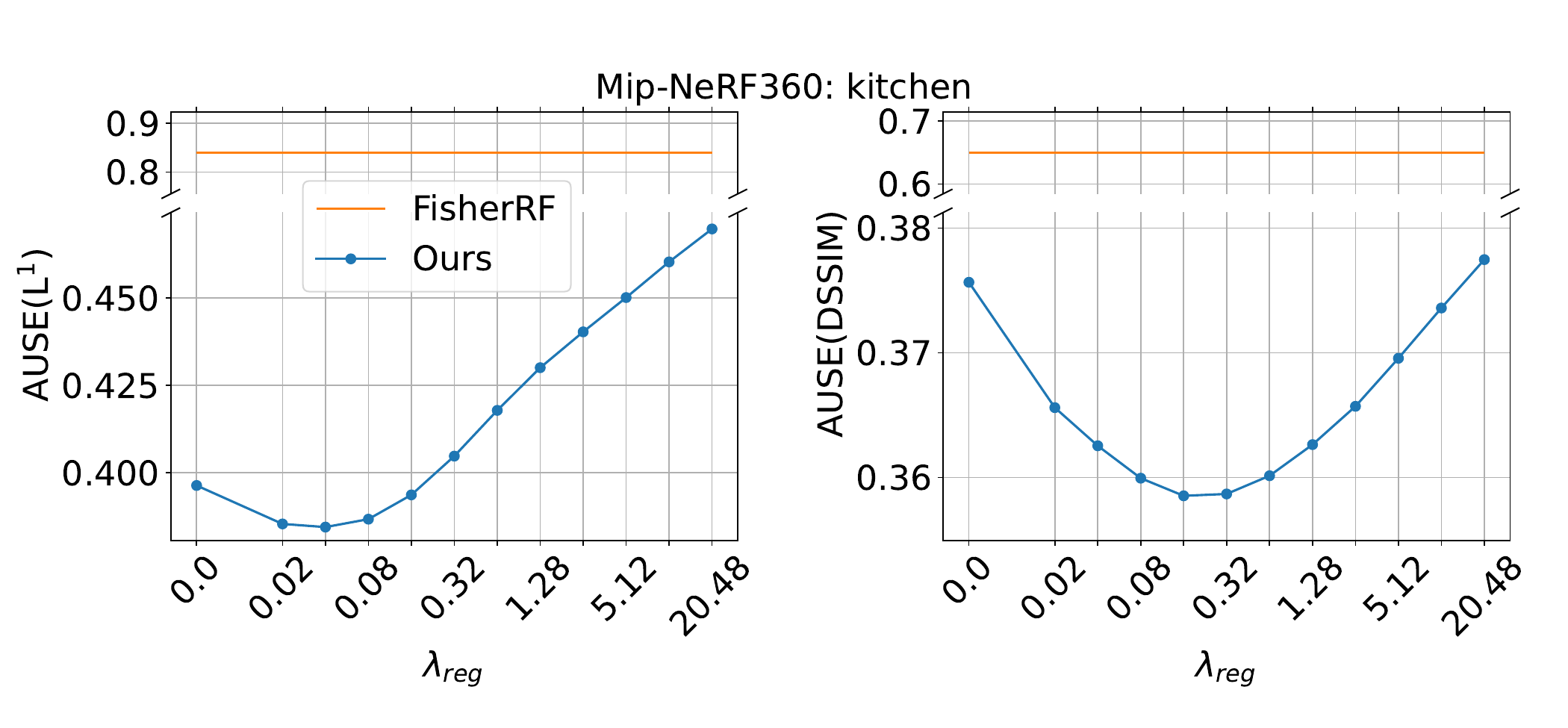}
    \caption{Scene-wise results for uncertainty estimation on highly novel views for the Mip-NeRF360~\citeNHL{barron2022mipnerf} scenes \textit{flowers}, \textit{garden}, and \textit{kitchen}.}
    \label{fig:sup:sparseUEscenesPlot2}
\end{figure}

\begin{figure}[t]
    \centering
    \includegraphics[width=\linewidth, trim=0.8cm 0.9cm 1.1cm 1.5cm, clip]{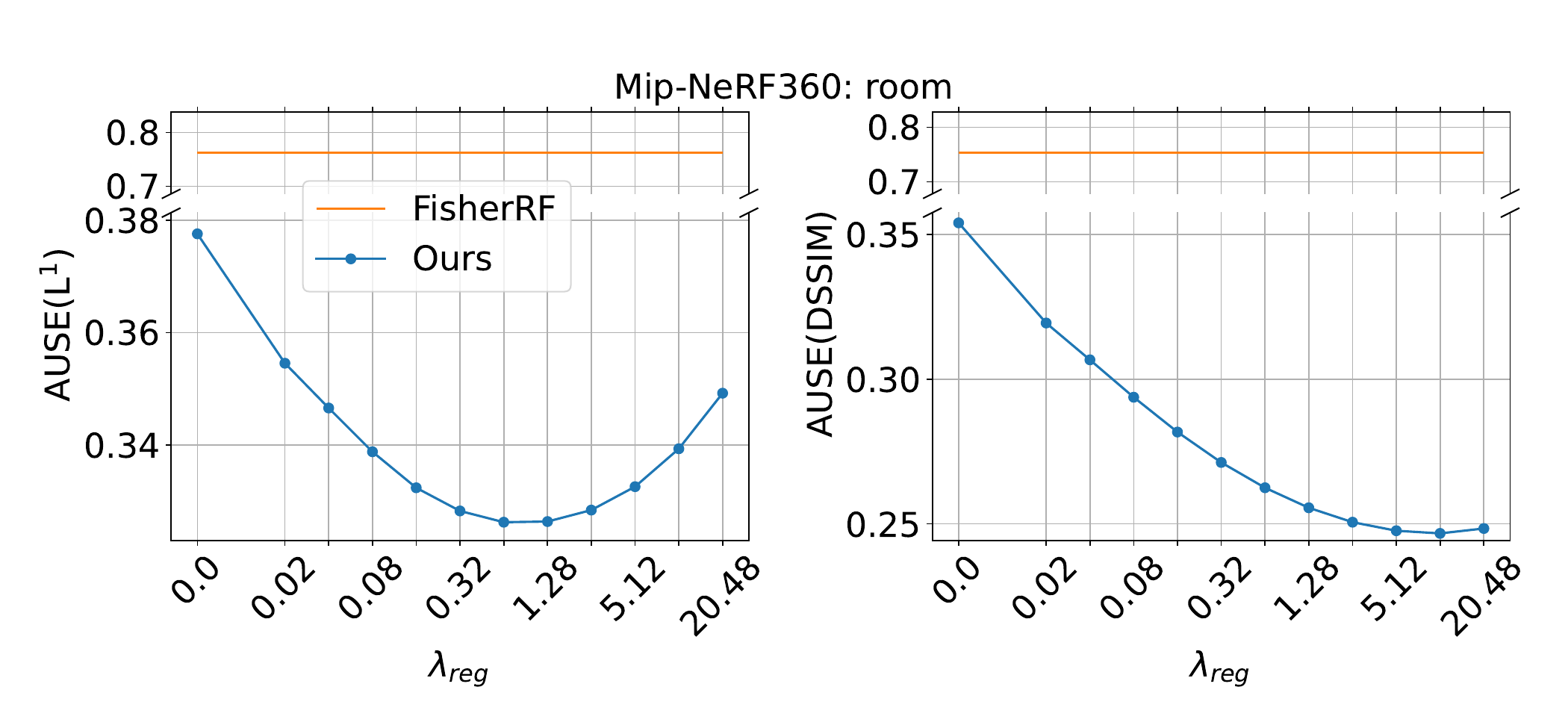}
    \includegraphics[width=\linewidth, trim=0.8cm 0.9cm 1.1cm 1.5cm, clip]{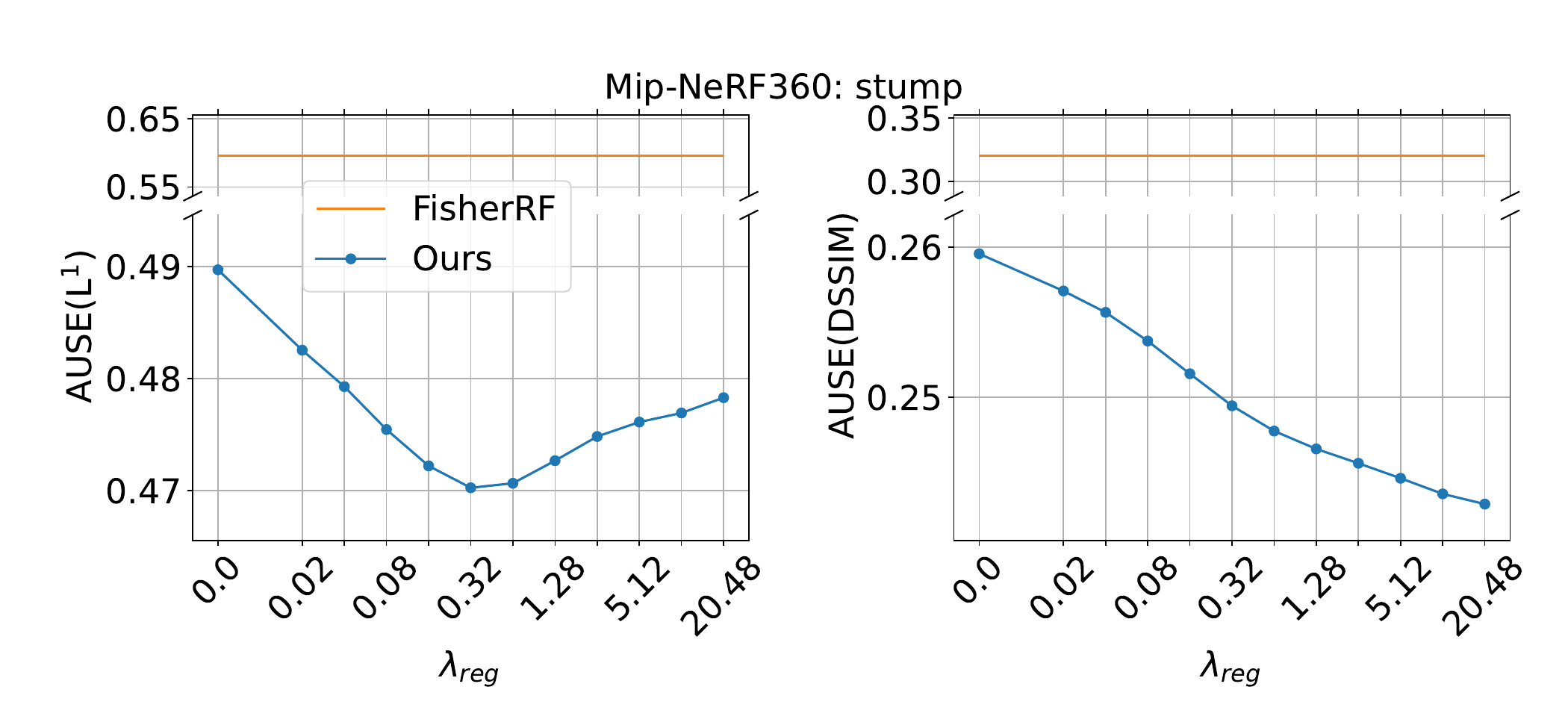}
    \includegraphics[width=\linewidth, trim=0.8cm 0.9cm 1.1cm 1.5cm, clip]{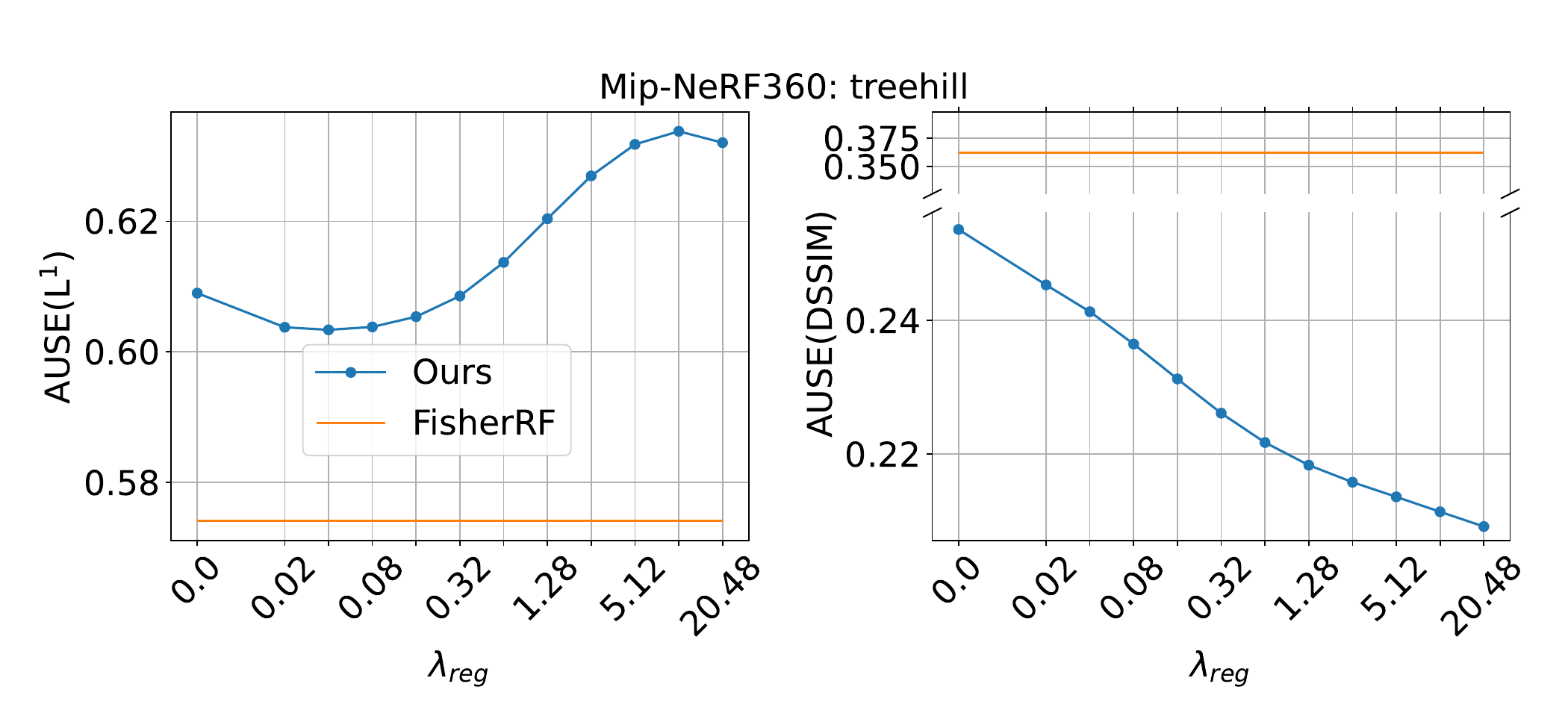}
    \caption{Scene-wise results for uncertainty estimation on highly novel views for the Mip-NeRF360~\citeNHL{barron2022mipnerf} scenes \textit{room}, \textit{stump}, and \textit{treehill}.}
    \label{fig:sup:sparseUEscenesPlot3}
\end{figure}

In \cref{fig:sup:sparseUEscenesPlot1,fig:sup:sparseUEscenesPlot2,fig:sup:sparseUEscenesPlot3}, we provide scene-wise results for our uncertainty estimation (UE) study on highly novel views, investigating the effectiveness of the proposed Bayesian-inspired regularization.
Overall, we observe consistent behavior across scenes. For the $L^1$ error metric, the lowest AUSE values are typically obtained for $0.08 \leq \lambda_{\text{reg}} \leq 0.32$, while for DSSIM the best performance is generally achieved for $\lambda_{\text{reg}} \geq 10.24$.

Notable exceptions are the \textit{bonsai}, \textit{kitchen}, and \textit{treehill} scenes.
For \textit{bonsai}, the $L^1$ AUSE exhibits a local minimum at $\lambda_{\text{reg}} = 0.04$, whereas the global minimum is attained at $\lambda_{\text{reg}} = 20.48$, the largest value considered in our study.
For \textit{kitchen}, DSSIM shows a pronounced minimum in the range $0.16 \leq \lambda_{\text{reg}} \leq 0.32$.
Finally, \textit{treehill} is the only scene where FisherRF slightly outperforms our method with respect to $L^1$ AUSE.
On all remaining scenes, our approach achieves lower AUSE values for both $L^1$ and DSSIM, often by a substantial margin.

\section{Splatting Quality Comparison}\label{sec:sup:splatQuality}

\begin{figure}
    \centering
    \includegraphics[width=0.98\linewidth]{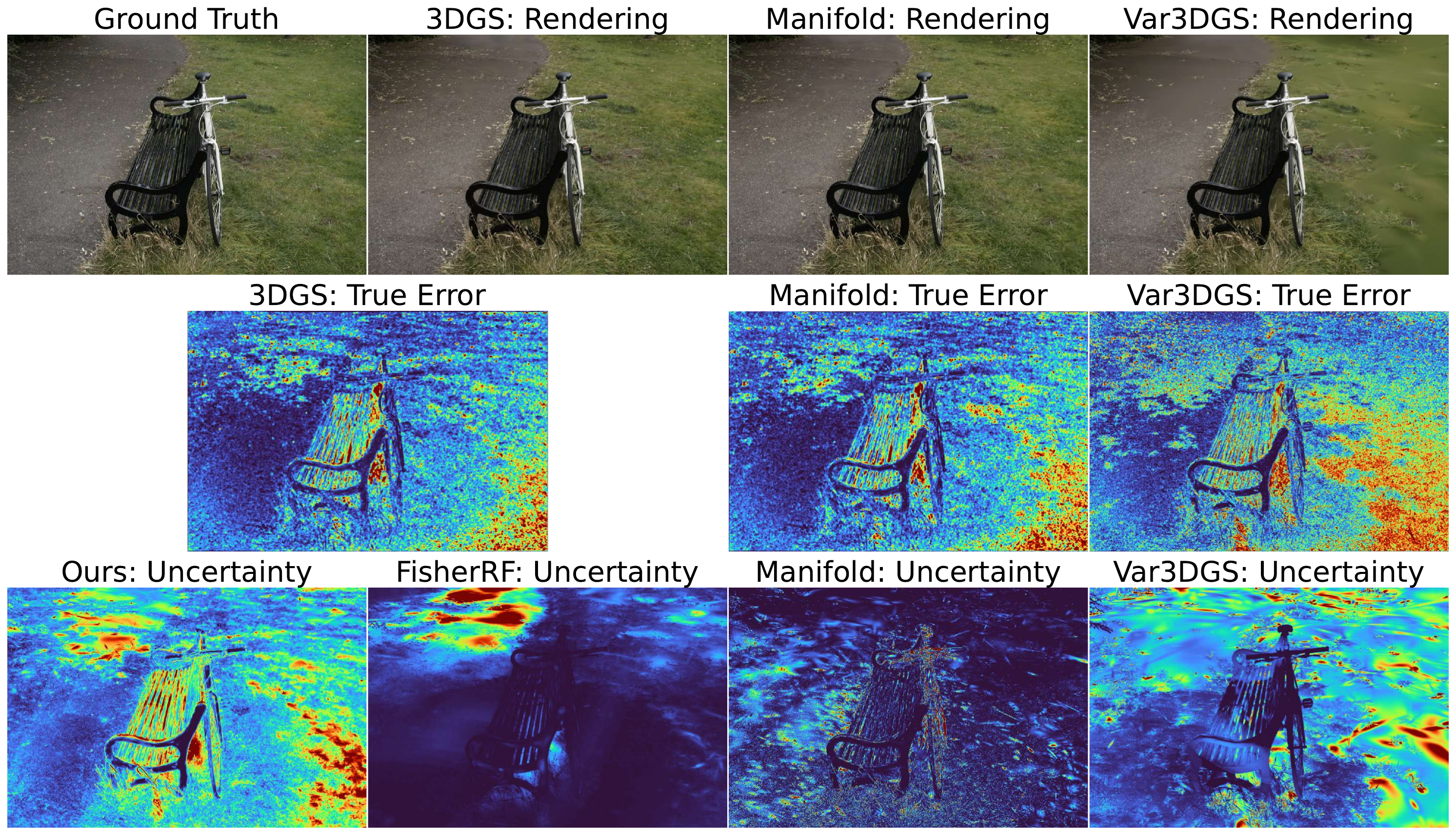}
    \includegraphics[width=0.98\linewidth]{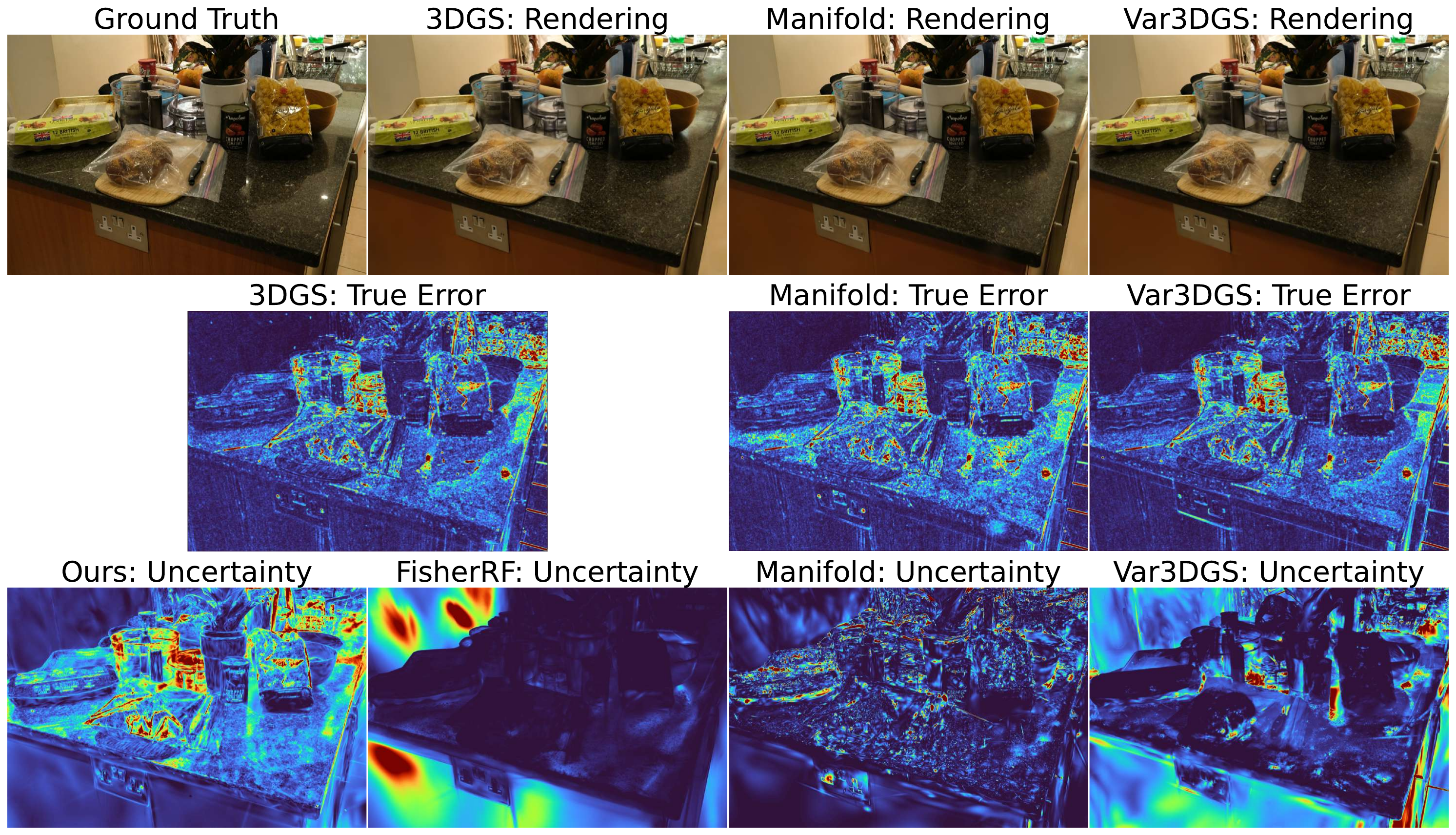}
    \caption{Qualitative comparison of predicted uncertainty maps to DSSIM error map. Including separate renderings and error maps for the stochastic 3DGS methods Manifold~\citeNHL{lyu2024manifold} and Var3DGS~\citeNHL{li2024variational}. Top: \textit{bicycle} from Mip-NeRF360~\citeNHL{barron2022mipnerf}. Bottom: \textit{counter} from Mip-NeRF360~\citeNHL{barron2022mipnerf}.}
    \label{fig:sup:UEqualitative1}
\end{figure}

\begin{figure}
    \centering
    \includegraphics[width=0.98\linewidth]{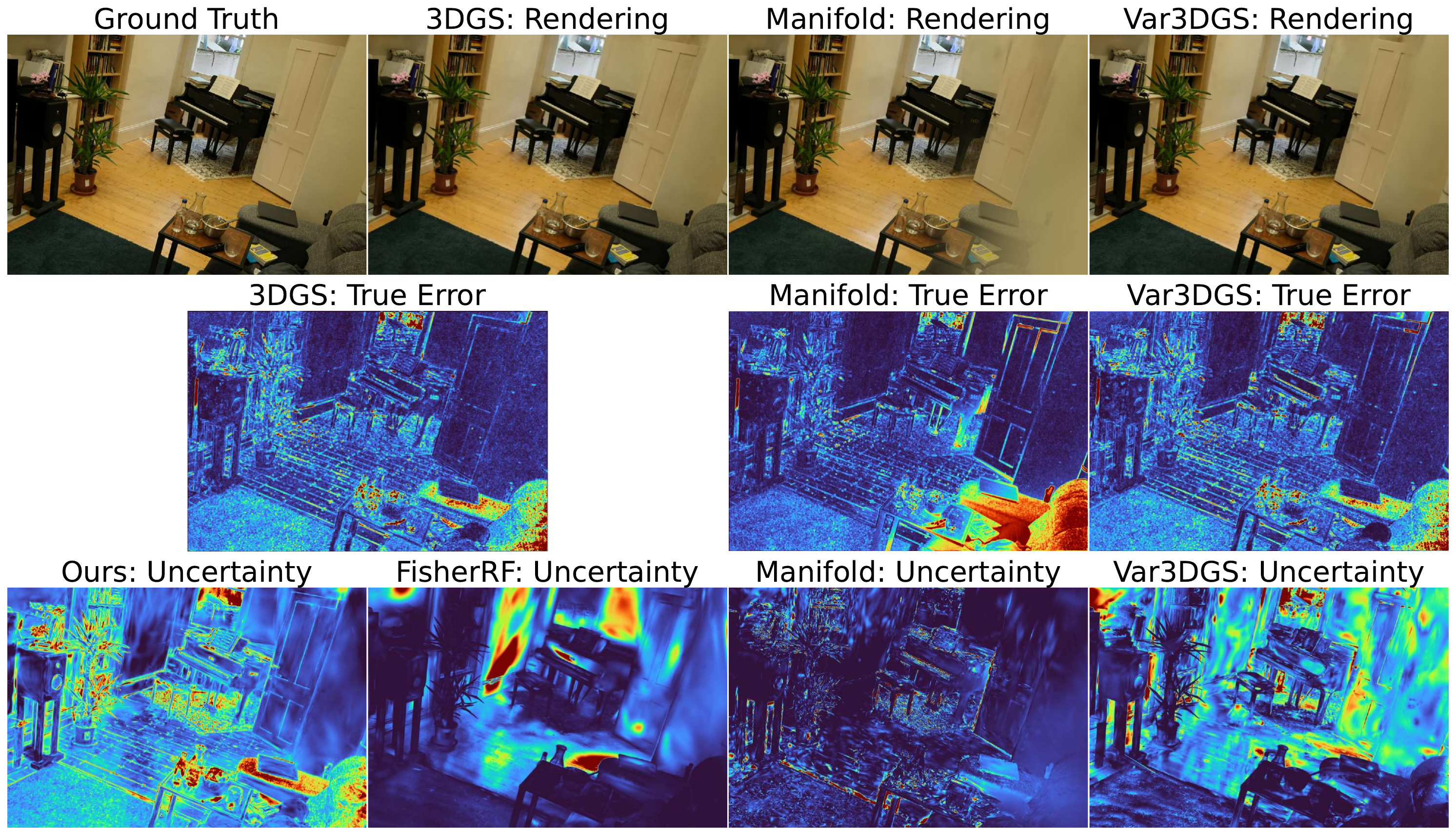}
    \includegraphics[width=0.98\linewidth]{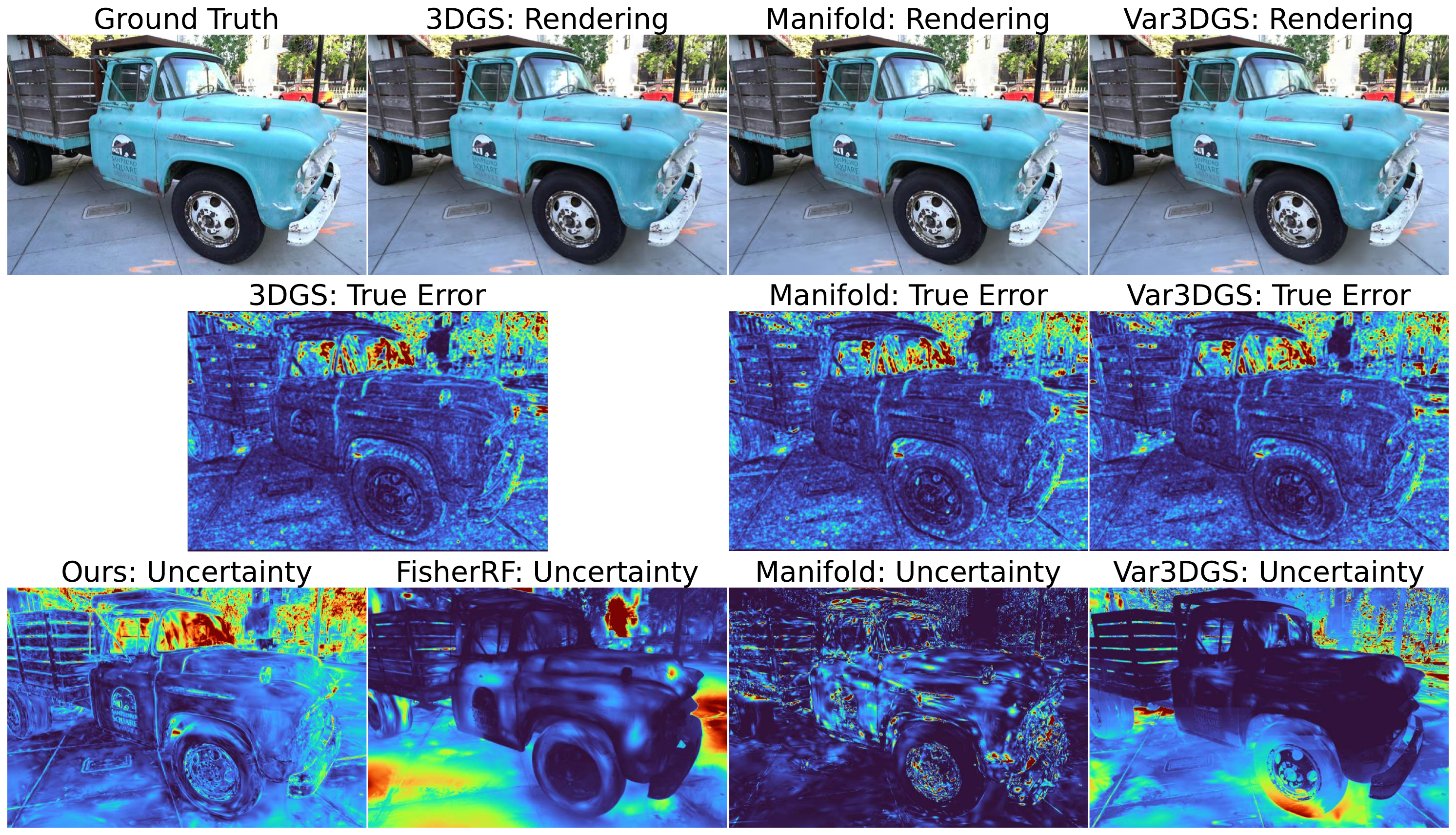}
    \caption{Qualitative comparison of predicted uncertainty maps to DSSIM error map. Including separate renderings and error maps for the stochastic 3DGS methods Manifold~\citeNHL{lyu2024manifold} and Var3DGS~\citeNHL{li2024variational}. Top: \textit{room} from Mip-NeRF360~\citeNHL{barron2022mipnerf}. Bottom: \textit{truck} from Tanks\&Temples~\citeNHL{knapitsch2017tanksandtemples}.}
    \label{fig:sup:UEqualitative2}
\end{figure}

In this section we compare the reconstruction quality of standard 3DGS~\citeNHL{kerbl20233dgaussiansplatting} against stochastic variants by Lyu et al.~\citeNHL{lyu2024manifold} (Manifold) and Li et al.~\citeNHL{li2024variational} (Var3DGS).
As detailed in \cref{tab:sup:quality}, standard 3DGS yields the highest reconstruction fidelity across nearly all metrics on our UE experimental datasets. The sole exception occurs in the two Tanks \& Temples scenes, where Manifold achieves a marginally higher PSNR but remains inferior in the other two quality metrics.
Qualitatively (\cref{fig:sup:UEqualitative1,fig:sup:UEqualitative2}), standard 3DGS also produces sharper details and fewer artifacts. For instance, Var3DGS yields very blurry grass in the \textit{bicycle} scene (see \cref{fig:sup:UEqualitative1}), while Manifold introduces an obstructing artifact in the \textit{room} scene (see \cref{fig:sup:UEqualitative2}). Despite these qualitative differences, the average rendering quality of all methods is similar.

\FloatBarrier

\begin{table}[b]
    \centering
    \caption{Quantitative comparison of reconstruction fidelity for different numbers of iterations per view $I$.}
    \label{tab:avs_it_comparision}
    \setlength{\tabcolsep}{3pt}
    \scalebox{0.75}{
    \begin{tabular}{rrrr}
        \toprule
        Iterations $I$ & PSNR$\ouruparrow$ & SSIM$\ouruparrow$ & LPIPS$\ourdownarrow$ \\
        \midrule
        10 & 20.316 & 0.617 & \tabhl{0.342} \\
        25 & 20.541 & \tabhl{0.618} & 0.343 \\
        50 & 20.676 & 0.615 & 0.344 \\
        100 & 20.604 & 0.612 & 0.346 \\
        200 & \tabhl{20.719} & 0.617 & 0.343 \\
        \bottomrule
    \end{tabular}}
\end{table}

\section{Iterations Prior to View Selection in AVS}\label{sec:sup:avs_training_iterations}

In our AVS framework, the uncertainty channels must be optimized prior to selecting the next best view.
We conduct experiments to determine the number of iterations required to achieve competitive performance. Following the practice of Jiang \etal~\citeNHL{jiang2024fisherrf}, we scale the number of iterations used to optimize the Gaussian primitive's uncertainty channels between selections with the current number of training views $N_\mathrm{views}$. Specifically, we train the uncertainty channels for $I \cdot N_\mathrm{views}$ iterations, where $I$ denotes the number of iterations per view.
\Cref{tab:avs_it_comparision} reports the reconstruction fidelity for several choices of $I$. No single value consistently yields the best performance across all metrics, and all results fall within a comparable range. For our main experiments, we choose $I = 50$, as it provides a substantial performance margin over all baselines while keeping the computational cost modest.
Although fewer iterations may lead to less accurate uncertainty heatmaps, we find that the induced ranking of candidate views by total uncertainty remains largely stable, leading to no reduced performance in AVS. The reduced optimization budget may even help to mitigate overfitting when only a small set of training views is available.

\section{UE Qualitative Examples}\label{sec:sup:qualitativeexamples}

In \cref{fig:sup:UEqualitative1,fig:sup:UEqualitative2} we provide additional qualitative UE examples. 
Here, we also include the renderings and DSSIM error maps for the stochastic 3DGS methods Manifold~\citeNHL{lyu2024manifold} and Var3DGS~\citeNHL{li2024variational}. 
As these two methods train separate splattings, their renderings and therefore their errors slightly differ from standard 3DGS, which we used for our main UE results for the post-hoc methods (our method and FisherRF~\citeNHL{jiang2024fisherrf}). 
This slightly limits the comparability between the post-hoc and stochastic 3DGS methods, as well as between the stochastic 3DGS methods themselves. 
However, as discussed in Appendix~\ref{sec:sup:splatQuality}, the overall reconstruction capabilities of these 3DGS variants are close to standard 3DGS, and in \cref{fig:sup:UEqualitative1,fig:sup:UEqualitative2} we see that their DSSIM error maps appear quite similar.

When comparing the uncertainty maps of the different UE methods to the true DSSIM error, we notice that our method seems to be the only one that reliably captures it. 
The regions indicated as most uncertain in the FisherRF uncertainty maps most of the time do not correspond to regions of high DSSIM error. 
Uncertainty maps of Manifold contain many small highlights that also often do not correspond to the DSSIM error. 
The uncertainty maps of Var3DGS visually come the closest to our method in quality, but they lack detail and still contain large regions that differ substantially from the DSSIM error.

\begin{table}[t]
    \centering
    \caption{Predicted uncertainty level ($\Delta$\,unc.: increase in mean uncertainty on test views) relative to the standard 3DGS setting, for blur/noise added to a fraction [\%] of the training views, or for a reduced number of training views. Results for the \textit{bicycle} scene of Mip-NeRF360~\citeNHL{barron2022mipnerf}. The predicted uncertainty increases monotonically under both aleatoric (blur/noise) and epistemic (fewer views) perturbations.}
    \label{tab:sup:uncertainty_response}
    \setlength{\tabcolsep}{6pt}
    \scalebox{0.85}{
    \begin{tabular}{lrrrrrrr}
        \toprule
        & \multicolumn{2}{c}{Blur} & \multicolumn{2}{c}{Noise} & \multicolumn{3}{c}{Num.\ Views} \\
        \cmidrule(lr){2-3}\cmidrule(lr){4-5}\cmidrule(lr){6-8}
        & 5\% & 10\% & 5\% & 10\% & 64 & 16 & 4 \\
        \midrule
        $\Delta$\,unc. & 119\% & 129\% & 127\% & 136\% & 122\% & 266\% & 596\% \\
        \bottomrule
    \end{tabular}}
\end{table}

\begin{figure}[t]
    \centering
    \includegraphics[width=0.98\linewidth]{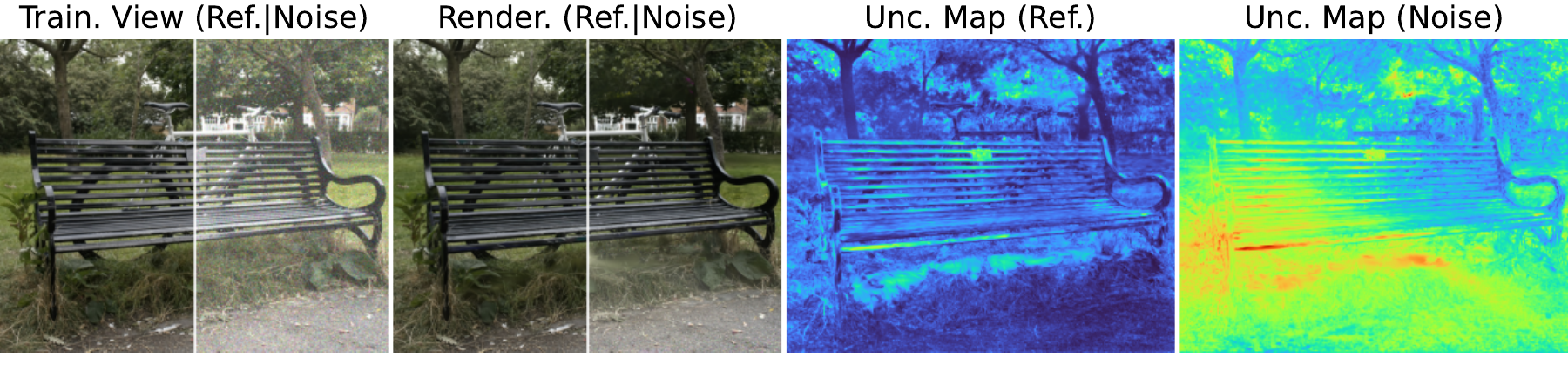}
    \caption{Qualitative example showing that adding Gaussian noise to the 3DGS training views (\textit{Noise}) increases the estimated uncertainty compared to 3DGS training without noise (\textit{Ref.}).}
    \label{fig:sup:noise_qualitative}
\end{figure}

\section{Uncertainty Response to Aleatoric and Epistemic Perturbations}\label{sec:sup:uncertainty_response}

Our method treats reconstruction residuals as a proxy for predictive uncertainty, aggregating aleatoric, epistemic, and optimization-related effects into a single signal (cf.\ \cref*{sec:un}).
To verify that this signal behaves as expected, we conduct a controlled study in which we inject \emph{aleatoric} uncertainty by corrupting a fraction of the training views with Gaussian blur or additive Gaussian noise, and \emph{epistemic} uncertainty by reducing the number of available training views.
For each setting, we report $\Delta$\,unc., the relative increase in mean predicted uncertainty on the held-out test views with respect to the standard (uncorrupted, full-view) 3DGS setting.
Experiments are performed on the \textit{bicycle} scene of Mip-NeRF360~\citeNHL{barron2022mipnerf}.

As shown in \cref{tab:sup:uncertainty_response}, the predicted uncertainty increases monotonically as we increase the fraction of corrupted training views (blur/noise) and as we reduce the number of training views.
This indicates that both aleatoric effects (introduced by image degradation) and epistemic effects (introduced by reduced view coverage) are captured and reflected in our predictive uncertainty estimates.
\cref{fig:sup:noise_qualitative} provides a qualitative example: adding Gaussian noise to the 3DGS training views (\textit{Noise}) visibly increases the estimated uncertainty compared to training without noise (\textit{Ref.}), while the corresponding renderings remain similar.


\end{document}